%% file: main.tex
\PassOptionsToPackage{numbers}{natbib}
\PassOptionsToPackage{table}{xcolor}
\documentclass{article}

\usepackage[utf8]{inputenc} % allow utf-8 input
\usepackage[T1]{fontenc}    % use 8-bit T1 fonts
\usepackage{hyperref}       % hyperlinks
\usepackage{url}            % simple URL typesetting
\usepackage{booktabs}       % professional-quality tables
\usepackage{amsfonts}       % blackboard math symbols
\usepackage{nicefrac}       % compact symbols for 1/2, etc.
\usepackage{microtype}      % microtypography
\usepackage{xcolor}         % colors

\usepackage[main, final, nonatbib]{neurips_2026}
\usepackage[numbers, sort&compress]{natbib}

\input{preamble}
\input{math_commands.tex}

\usepackage{eccvabbrv}

\usepackage{graphicx}
\usepackage{booktabs}

\usepackage[accsupp]{axessibility}  % Improves PDF readability for those with disabilities.

\makeatletter
\newcommand{\instfootnote}[1]{%
    \begingroup
    \renewcommand{\@makefntext}[1]{\noindent##1}% 
    \renewcommand{\thefootnote}{}%
    \footnotetext{#1}%
    \endgroup
}
\makeatother

\begin{document}

\title{Structured State-Space Regularization for \\
Generation-Friendly Image Tokenization} 

\author{
Jinsung Lee$^1$\\
{\tt\small jinsunglee@postech.ac.kr}
\And
Jaemin Oh$^2$\\
{\tt\small jaemin\_oh@brown.edu}
\And
Namhun Kim$^1$\\
{\tt\small namhunkim@postech.ac.kr}
\And
{Dongwon Kim$^3$} \\
{\ \ \ \quad \tt\small kdwon@kaist.ac.kr\qquad} \\
\And
Byung-Jun Yoon$^{4,5}$\\
{\tt\small bjyoon@tamu.edu}\\
\And
Suha Kwak$^1$\\
{\tt\small suha.kwak@postech.ac.kr}
\And
\mbox{\href{https://jinsingsangsung.github.io/posts/structured-state-space-regularization-for-generation-friendly-image-tokenization/}{\textcolor[HTML]{38669F}{\texttt{https://jinsingsangsung.github.io/WHippo}}}}
}
\instfootnote{
\begin{tabular}{ccccc}
{$^1$ POSTECH} &
{$^2$ Brown University} &
{$^3$ KAIST} & 
{$^4$ Texas A\&M University} &
{$^5$ Brookhaven National Laboratory}
\end{tabular}
}
\maketitle

\input{sections/0_abstract}
\input{sections/1_intro}
\input{sections/3_preliminaries}
\input{sections/4_method}
\input{sections/5_experiments}
\input{sections/6_conclusion}

% \clearpage  
\vspace{4mm}

% \section*{Acknowledgements}
% Please insert your acknowledgments here.


\bibliographystyle{splncs04}
\bibliography{main}

\input{sections/10_supplementary}
% \clearpage  
% \input{sections/7_paper-checklist}
\end{document}

%% file: preamble.tex
\def\ie{\emph{i.e.}}
\def\eg{\emph{e.g.}}
\def\etal{\emph{et al.}}

\usepackage{graphicx, subcaption, adjustbox}
\usepackage{makecell}
\usepackage{pgf}
\usepackage{booktabs, multirow, multicol}
\usepackage{makecell}
\usepackage{colortbl}
\usepackage{enumitem}
\usepackage{soul}
\usepackage[table]{xcolor}
\usepackage{xfp}

\usepackage[subrefformat=simple,labelformat=simple,font=footnotesize]{subcaption}

\usepackage{amsmath, amssymb, mathtools, amsthm}

\newcommand{\overbar}[1]{\mkern 1.5mu\overline{\mkern-1.5mu#1\mkern-1.5mu}\mkern 1.5mu}

\newtheorem{prop}{Proposition}[section]

\newcommand{\hmmax}{110}  % value mapped to full color
\newcommand{\hmtop}{65}   % max color intensity (%)
\newcommand{\heat}[1]{
  \cellcolor{blue!\fpeval{round(min(\hmtop, \hmtop*#1/\hmmax))}}#1}

\newcommand{\rel}[1]{\,{\footnotesize\textcolor{gray!80!black}{#1\%}}}
\newcommand{\norel}{\phantom{\rel{+0.0}}}

%% file: math_commands.tex
\usepackage{amsmath,amsfonts,bm}

\def\eqref#1{equation~\ref{#1}}
\def\1{\bm{1}}

\def\rvx{{\mathbf{x}}}

\def\rmA{{\mathbf{A}}}
\def\rmB{{\mathbf{B}}}
\def\rmC{{\mathbf{C}}}

\def\rmI{{\mathbf{I}}}

\DeclareMathAlphabet{\mathsfit}{\encodingdefault}{\sfdefault}{m}{sl}
\SetMathAlphabet{\mathsfit}{bold}{\encodingdefault}{\sfdefault}{bx}{n}

\def\sR{{\mathbb{R}}}

%% file: sections/0_abstract.tex
\begin{abstract}
\label{sec:abstract}
Image tokenizers play a central role in modern generative models, where the structure of the latent space critically determines the downstream generation performance.
A key but underexplored property of effective latent representations is \textit{spectral organization}, the ability to encode information across frequency components.
In this work, we introduce \textit{structured state-space regularization}, a principled approach to inducing spectral structure in latent spaces.
We derive a regularization objective by revisiting state-space models (SSMs) as systems mimicking a basis function's behavior.
This perspective reveals that hidden states of SSMs are induced to capture the frequency components, resulting in a novel regularizer
that enforces the latent space to capture spectral structure of images.
Experiments demonstrate that our regularizer improves the generative performance of image tokenizers while incurring only minimal loss in their reconstruction fidelity.
\end{abstract}

%% file: sections/1_intro.tex
\section{Introduction}
\label{sec:intro}

Recent advances in computer vision increasingly rely on image tokenizers.
Generative models~\cite{rombach2022high, vahdat2021score, tian2024visual, li2024autoregressive}, world models~\cite{NWM, CosmosTokenizer, hafner2019dream}, and representation models~\cite{JEPA, bardes2024revisiting} 
are all moving toward paradigms where modeling is performed in a learned latent space rather than in pixel space.
This shift is driven by several practical advantages: latent representations are lower-dimensional, enabling substantially cheaper computation than high-dimensional pixel representation, and are able to capture  
semantic layout that raw pixel intensities do not explicitly encode.
As a result, image tokenizers, which shape the geometry and content of the latent space, have become a critical component in vision models, and improving them is of significant value as it can directly boost the efficiency and quality of modern generative AI systems.

A growing body of recent work focus on constructing latent spaces that are \textit{generation-friendly}; meaning that they can be modeled accurately and efficiently by downstream  generative models.
Such a property is crucial for applications that rely on probabilistic sampling, including planning~\cite{kim2026planning, NWM}, conditional generation~\cite{van2016conditional, ho2022classifier}, editing~\cite{kim2022diffusionclip, kawar2023imagic}, and manipulation~\cite{black2024pi_0, ye2026world}.
Despite its importance, the precise characteristics that make a latent space generation-friendly remain unclear and are actively being investigated.
Prior studies suggest several contributing factors: Skorokhodov~\etal~\cite{skorokhodov2025improving} and Kouzelis~\etal~\cite{EQVAE} highlight the role of scale and rotation equivariance between the pixel and latent spaces, while Chen~\etal~\cite{chen2025masked} insist the number of Gaussian mixture modes required to approximate latent distribution as a key factor.
More recently, incorporating features from large vision foundation models~\cite{vavae,repae,RAE} has emerged as a highly effective approach for improving generation-friendliness.

In this work, we argue that an important yet largely overlooked property underlying generation-friendly latent spaces is \textit{spectral organization}.
Viewing image generation as an ill-posed inverse problem~\cite{chung2022diffusion, chung2022improving}, exploiting the frequency structure of natural images has long been recognized as a powerful inductive bias~\cite{phillips2022spectral, vandewalle2006frequency, pan1999two}.
Natural images exhibit characteristic spectral statistics, such as power-law distributions over frequencies~\cite{castleman1979digital}, which have been successfully utilized in classical and modern approaches to image restoration~\cite{hummel1985gaussian,  fu2021dw, jiang2024fast} and compression~\cite{DCT, wallace1991jpeg}.
Surprisingly, however, the role of such spectral structure has received limited attention in the design of latent spaces for generative modeling.
A few prior attempts~\cite{CosmosTokenizer, ning2024dctdiff, esteves2025spectral} incorporate frequency-domain representations directly into the tokenizer, but these approaches often suffer from significantly different training dynamics and achieve suboptimal performance due to the discrepancy between frequency and spatial domains.

To address this limitation, we introduce spectral structure into latent spaces in a principled yet non-disruptive manner by drawing inspiration from state-space models (SSMs).
SSMs implicitly encode an input signal using a set of basis functions whose coefficients correspond to the spectral components of the signal~\cite{gu2022train, yu2024tuning, lee2026exploring}.
In particular, the hidden state of SSMs is trained to react to a signal transformation in a way that the basis function's coefficients would behave, thereby implicitly equipping the capability to capture the spectral structure of the input signal.
Building on this insight, we introduce \textbf{structured state-space regularization}, a lightweight objective that encourages latent features to achieve spectral organization without altering the underlying tokenizer architecture.
Our framework not only enhances the image tokenizer's generation-friendliness with minimal loss in reconstruction fidelity, but also provides intriguing insights involving SSMs as a promising, yet largely unexplored, modeling tool for image tokenization.

In summary, our contribution is three-fold:
\begin{itemize}[leftmargin=0.5cm]
    \item We propose a novel, theoretically motivated regularizer to enhance image tokenizer's downstream generation performance.
    \item We present the first theoretical framework for incorporating principles of SSMs into image tokenization.
    \item Our method proves effective on widely used image tokenizers, improving generation performance with minimal sacrifice in reconstruction quality.
\end{itemize}

%% file: sections/3_preliminaries.tex
\section{Preliminaries}
\label{sec:preliminaries}

As our method is formulated within the state-space model (SSM) framework, we first introduce the notation of SSMs, describe their operational principles, and explain their practical implementation.

\subsection{SSMs}
In the context of machine learning, SSMs are sequence-to-sequence models that map the input function $u(t)$ to the output $y(t)$, employing the state-space representation of a linear system defined as follows: \\
\begin{minipage}{0.5\linewidth}
    \begin{align}\label{eq:ssm_cont1}
        \frac{\mathrm{d}}{\mathrm{d}t}\rvx(t) = \rmA \rvx(t) + \rmB u(t),
    \end{align}
\end{minipage}
\quad
\begin{minipage}{0.35\linewidth}
    \begin{align}\label{eq:ssm_cont2}
        y(t) = \rmC \rvx(t),
    \end{align}
\end{minipage}\vspace{2.0mm}\\
where $\rvx(t) \in \sR^{N \times 1}$, $u(t) \in \sR$, $y(t) \in \sR$,  $\rmA \in \sR^{N \times N}$, $\rmB \in \sR^{N \times 1}$, and $\rmC \in \sR^{1\times N}$.
Eq.~(\ref{eq:ssm_cont1}) specifies the direction in which the hidden state $\rvx(t)$ is updated to incorporate the input $u(t)$ at time $t$, and 
Eq.~(\ref{eq:ssm_cont2})
produces the output $y(t)$ from the hidden state $\rvx(t)$ using linear projection, similar to the linear state-space system from control theory~\cite{zadeh2008linear} or language models~\cite{radford2018improving}.
Here, the state transition matrix $\rmA$ is particularly important, as it determines the way the hidden state evolves to integrate new inputs over time.
HiPPO matrix~\cite{gu2020hippo} is a family of the state transition matrix $\rmA$ that enhances the long-term dependency of the memory $\rvx$, which is derived from projecting the input $u(t)$ to predefined orthogonal polynomial bases $\{\phi_n(t)\}^{N}_{n=1}$ to obtain their coefficients $\{c_n(t)\}^{N}_{n=1}$ as follows:
\begin{equation}\label{eq:coef_hippo}
    c_n(t) =\langle {u_{\leq t}}, \phi_n \rangle_{\nu} = \int_{0}^{t} u(x) \phi_n(x)
    \nu(x)~\mathrm{d}x,
\end{equation}
where $\nu$ is a measure depending on the choice of bases.
Here, the HiPPO matrix $\mathbf{A}$ is derived to describe the evolution of $c_n(t)$: as the streaming input $u_{\leq t}$ accumulates over time, its basis projection $c_n(t)$ changes continuously, and $\mathbf{A}$ can be deterministically obtained by differentiating $c_n(t)$ with respect to $t$,
yielding a differential equation that governs its dynamics.
Numerous studies have shown that the structure and properties of HiPPO matrices enhance the performance of SSMs: \eg, the sparse structure of $\rmA$ enables efficient matrix-vector multiplication in Eq.~(\ref{eq:ssm_cont1}) and initializing $\rmA$ with a HiPPO matrix yields superior performance compared to random initialization~\cite{gu2021combining, gu2021efficiently, gu2022parameterization, gu2022train, hasaniliquid}.

\subsection{Discretization}
Since observations are inherently discrete samples (\eg, images can be regarded as pixels sampled from an underlying continuous signal), the continuous-time dynamics in Eq.~(\ref{eq:ssm_cont1}) must be approximated with a step size parameter $\Delta \in \mathbb{R}_{+}$, yielding the recursive form:\\
\begin{minipage}{0.5\linewidth}
    \begin{align}\label{eq:ssm_disc1}
        \rvx_{t} = \overbar{\rmA} \rvx_{t-1} + \overbar{\rmB} u_{t}, 
    \end{align}
\end{minipage}
\quad
\begin{minipage}{0.35\linewidth}
    \begin{align}\label{eq:ssm_disc2}
        y_{t} = \rmC \rvx_{t},
    \end{align}
\end{minipage}\vspace{1.7mm}\\
where $\rvx_t := \rvx(t)$, $u_t := u(t)$, and $y_t := y(t)$ for $t\in \mathbb{N}$.
The discretized form admits an interpretation in which $\rvx_{t}$ represents the compressed history accumulated up to the $t$-th timestep, and the new information $u_t$ is integrated into the compressed state to update $\rvx_{t-1}$ to $\rvx_t$.
Common discretization methods include Euler, bilinear~\cite{gu2020hippo, gu2021combining, gu2021efficiently}, 
and zero-order hold (ZOH)~\cite{voelker2019legendre, gupta2022diagonal}.
For instance, applying the forward Euler discretization to Eq.~(\ref{eq:ssm_cont1}) results in:
\begin{equation}\label{eq:euler_disc_example}
\begin{aligned}
    \rvx(t_i) &\approx \rvx(t_{i-1}) + \Delta \frac{\mathrm{d}}{\mathrm{d}t}\rvx(t) \big|_{t=t_{i-1}} \quad (\because \textrm{Forward Euler}) \\
    &= \rvx(t_{i-1}) + \Delta \big( \rmA \rvx(t_{i-1}) + \rmB u(t_i) \big) \\
    &:= \overbar{\rmA}\rvx(t_{i-1}) + \overbar{\rmB}u(t_i).
\end{aligned}
\end{equation}
Through these methods, $\overbar{\mathbf{A}}$ and $\overbar{\mathbf{B}}$ are expressed with respect to $\mathbf{A}, \mathbf{B}$, and $\Delta$.

%% file: sections/4_method.tex
\section{Proposed method}
\label{sec:method}
In this section, we examine the mechanisms that enable SSMs to achieve spectral organization, and introduce a novel regularization method that leverages these mechanisms to regularize tokenizers towards a generation-friendly direction.

\subsection{A generalized view of SSMs: basis projection and input transformation}\label{sec:gSSM}
What truly constructs SSMs?
At their core, SSMs rely on a form of \textit{basis projection} that summarizes the input, while allowing the update of the compressed representation to reflect the \textit{input transformation}.
We argue that such an update represents the core principle of SSMs that leads to a notion of \textit{generalized view of state-space models}, which can be characterized by two key components:
\vspace{1mm}
\begin{enumerate}
    \item a \textit{basis projection} $\mathbf{c}:
    \bigcup\limits_{L\in \mathbb{N}_{\scriptscriptstyle{\geq N}}}
    \mathbb{R}^{L} \rightarrow \mathbb{R}^{N}$,
    \item an \textit{input transformation} $\theta : \mathbf{I}_{t-1} \mapsto \mathbf{I}_{t}$, where $\mathbf{I}_{t}$ denotes the input at time $t$.
\end{enumerate}
\vspace{1mm}
The choice of these two components directly determines the hidden state's update rule $\mathbf{x}_{t-1} \mapsto \mathbf{x}_{t}$, since the SSM formulation (Eq.~(\ref{eq:ssm_cont1})) adopts its update rule from the basis coefficient's dynamics computed on the predefined input transformation $\mathbf{c}(\mathbf{I}_{t-1}) \mapsto \mathbf{c}(\mathbf{I}_{t})$. 
For instance, the update rule $\mathbf{x}_{t-1} \mapsto \mathbf{x}_{t}$ in Eq.~(\ref{eq:ssm_disc1}) is adopted from the basis coefficients' update rule $\mathbf{c}(\mathbf{I}_{t-1}) \mapsto \mathbf{c}(\mathbf{I}_{t})$, 
where $\mathbf{c}(\cdot)$ is defined as a projection onto orthogonal polynomial bases
while $\theta: \mathbf{I}_{t-1} \mapsto \mathbf{I}_t$ represents the concatenation of the $t$-th token next to the 1D input at time $t-1$~\cite{gu2020hippo}.
If we let $\mathbf{c}(\cdot)$ be a projection onto 2D orthogonal polynomial bases and $\{\mathbf{I}_t\}$ be a corner-based sweeping construction of a 2D image, we obtain 2D SSMs proposed by Baron \etal~\cite{baron20242} and Nguyen \etal~\cite{nguyen2022s4nd}. 

This perspective on SSMs provides two key insights.
First, it provides an intuitive explanation for SSM's frequency-capturing property~\cite{gu2022train, lee2026exploring, yu2024tuning}.
Note that basis coefficients correspond to the magnitudes of spectral components in the input signal.
Since the hidden state $\mathbf{x}_t$ is trained to resemble the behavior of the basis coefficients $\mathbf{c}(\cdot)$, its ability to capture frequency information can be understood as arising from its resemblance to $\mathbf{c}(\cdot)$.
This is also connected to the well-known ability of SSMs to produce compact representations, as basis representations are widely regarded as a compact way to encode signals~\cite{wallace1991jpeg, DCT}.
Second, this view naturally extends SSMs beyond the standard formulation in Eq.~(\ref{eq:ssm_disc1}).
By selecting different pairs of basis projections and input transformations $\bigl(\mathbf{c}(\cdot), \theta(\cdot) \bigr)$ that yield tractable update rule $\mathbf{c}(\mathbf{I}_{t-1}) \mapsto \mathbf{c}(\mathbf{I}_{t})$, one can generalize SSMs to a broader range of modalities.
We illustrate this generalization in Fig.~\ref{fig:teaser}.

\input{figures/teaser}

\subsection{Structured state-space regularization}\label{sec:whippo}

Building on this philosophy, we propose \textbf{structured state-space regularization}, a novel regularizing formulation that emerges from the generalized view of the SSMs by extending the choice of the basis projection $\mathbf{c}(\cdot)$ and the input transformation $\theta(\cdot)$ that operates on 2D data (Fig.~\ref{fig:whippo}).

\vspace{-1mm}
\paragraph{Choice of the basis projection $\mathbf{c}(\cdot)$} \label{sec:compression_function}
A natural choice of $\mathbf{c}(\cdot)$ that allows update rules with favorable mathematical properties\footnote{
See Appendix~\ref{appendix:compression_functions} for details.
}  is a projection using 2D basis functions like Fourier bases or 2D orthogonal polynomials~\cite{gu2020hippo, gu2022train}.
Formally, let the 2D orthogonal basis functions be denoted as $\{\phi_n\}_{n\in\mathbb{N}}$.
We define the input image as a continuous surface function $\mathbf{I}(x,y): [0, W] \times [0, H] \rightarrow \mathbb{R}$,
where $\mathbf{I}(w,h)$ indicates the pixel intensity of the cell located at the $(w, h)$-th position.
We assume a single-channel image for simplicity, as the extension to multi-channel can be achieved by independently applying the same logic to each channel~\cite{gu2021combining}.
Let $c_n$ be the coefficient obtained by projecting the image $\mathbf{I}(x,y)$ onto the $n$-th orthonormal basis function $\phi_n$:
\begin{equation}
    c_n =\langle \mathbf{I}, \phi_n \rangle = \iint\limits_{[0,W]\times [0,H]} \mathbf{I}(x,y) \phi_n(x,y) ~ \mathrm{d}y \mathrm{d}x,
\end{equation}
then, by the Parseval's identity, the surface function $\mathbf{I}(x,y)$ is expressed as a linear combination of the basis functions $\{\phi_n\}_{n\in \mathbb{N}}$ with the coefficients $\{c_n\}_{n\in \mathbb{N}}$:
\begin{equation}\label{eq:image_approx}
    \mathbf{I}(x,y) = \sum^{\infty}_{n=1}c_n(\mathbf{I})\phi_n(x,y).
\end{equation}
If some terms in the expansion are omitted, the reconstruction becomes an approximation. 
Thus, we truncate the terms in Eq.~(\ref{eq:image_approx}) to define the basis projection as $\mathbf{c}(\mathbf{I}):= 
\begin{bmatrix}
    c_1 & c_2 & \cdots & c_N
\end{bmatrix}^{\top}$,
where $N \leq HW$.

\vspace{-1mm}
\paragraph{Choice of the input transformation $\theta(\cdot)$}
Among various explorations for transforming an image such as autoregressive modeling~\cite{lee2022autoregressive, ramesh2021zero, sun2024autoregressive}, random masking/unmasking~\cite{li2024autoregressive, chang2022maskgit, li2023mage, he2022masked}, pooling~\cite{tian2024visual}, or denoising~\cite{ho2020denoising, sohl2015deep}, we adopt a \textit{progressive blurring process based on Gaussian blur}~\cite{ballard1982computer} for two main reasons.
First, Gaussian blur offers favorable mathematical properties, including linearity and differentiability, while its derivative being conveniently derived from the fundamental solution of the heat equation~\cite{perona2002scale, weickert1998anisotropic}.
Second, it aligns well with the inductive bias of images: Gaussian blur is shift-invariant, preserves existing structures without introducing artifacts~\cite{lindeberg1994scale}.
We examine both blurring and deblurring transformations and find that the blurring process emerges as the more natural choice during derivation.
The details for its derivation and justification are provided in Appendix~\ref{appendix:deblurring}.
Accordingly, we present the derivation based on the blurring process.
Formally, given an image $\mathbf{I} = \mathbf{I}_0$, we define the progressive Gaussian blurring process $\{\mathbf{I}_t \}_{t=0}^{T}$ as a series of uniformly sampled images from a continuous process $\{\mathbf{I}_\tau\}_{\tau \in [0,T \Delta]}$ by letting $\mathbf{I}_t := \mathbf{I}_{\tau}|_{\tau = t \Delta}$, where $\Delta$ is the step size between adjacent samples.
For $\tau \in (0, T]$, 
\begin{align}\label{eq:gaussian_convolution}
    \mathbf{I}_{\tau} = G_{\tau} * \mathbf{I},
\end{align}
where $G_{\sigma^2}(x,y) = \frac{1}{2\pi \sigma^2}\exp(-\frac{x^2 + y^2}{2\sigma^2})$ denotes the Gaussian blur kernel.

\vspace{-1mm}
\paragraph{Update rule $\mathbf{c}(\mathbf{I}_{t-1}) \mapsto \mathbf{c}(\mathbf{I}_t)$} Owing to the well-studied analytical properties of Gaussian blur, we can formulate how $\mathbf{c}(\mathbf{I}_t)$ evolves as $t$ increases, by deriving the derivatives of $\mathbf{c}(\mathbf{I}_\tau)$ with respect to $\tau$.
For brevity, we demonstrate the detailed derivation process in Appendix~\ref{appendix:update_derivations}, and present only the resulting formulation below.

\begin{prop}
Let $\mathbf{c}(\cdot)$ be a projection using 2D orthogonal basis functions $\{\phi_k\}_{k=1}^{N}$ and $\{\mathbf{I}_\tau\}$ be a continuous Gaussian blurring process.
Then, the derivative $\frac{\mathrm{d}}{\mathrm{d}\tau}\mathbf{c}(\mathbf{I}_\tau)$ and the update rule $\mathbf{c}(\mathbf{I}_{t-1}) \mapsto \mathbf{c}(\mathbf{I}_t)$ are derived as follows:
    \begin{align}
        \frac{\mathrm{d}}{\mathrm{d}\tau}\mathbf{c}(\mathbf{I}_\tau) 
        &= \mathbf{A} \mathbf{c}(\mathbf{I}_\tau), \qquad \textrm{where} \ \ 
        \mathbf{A} = \frac{1}{2}
        \begin{bmatrix}
            \langle  \nabla^2\phi_1, \phi_1 \rangle & \cdots & \langle \nabla^2\phi_{N}, \phi_{1} \rangle \\
            \langle  \nabla^2\phi_1, \phi_2 \rangle & \cdots & \langle \nabla^2\phi_{N}, \phi_{2} \rangle \\
            \vdots & \ddots & \vdots \\
            \langle  \nabla^2\phi_{1}, \phi_{N} \rangle & \cdots & \langle  \nabla^2\phi_{N}, \phi_{N} \rangle
        \end{bmatrix} ,\label{eq:whippo_dynamics} \\
        \mathbf{c}(\mathbf{I}_t) &= \bar{\mathbf{A}}\mathbf{c} (\mathbf{I}_{t-1}), \qquad \ \textrm{where} \ \ 
        \bar{\mathbf{A}} = \delta(\Delta, \mathbf{A}) \ \textrm{ for a discretization method} \ \delta.
        \label{eq:whippo_update}
    \end{align}
\end{prop}

\paragraph{Structured state-space regularization}
From the update rule in Eq.~(\ref{eq:whippo_update}), one can naturally derive a regularizing term for a trainable encoder network $\mathcal{E}: \rmI \mapsto \mathcal{E}(\rmI)$ that maps an image to a latent.
Recall that the hidden state of SSMs obtains basis function properties by following the coefficient dynamics $\frac{\mathrm{d}}{\mathrm{d}t}\mathbf{c}(\rmI_{t})$, or equivalently, the discretized update rule $\mathbf{c}(\rmI_{t-1}) \mapsto \mathbf{c}(\rmI_{t})$ (Sec.~\ref{sec:gSSM})).
Since we aim to endow basis-like inductive bias to the encoder output $\mathcal{E}(\rmI)$, we let $\mathcal{E}$ follow the derived update rule over the predefined transformation $\{\rmI_t\}$:
\begin{equation}\label{eq:whippo_update2}
    \mathcal{E}(\rmI_t) = \bar{\rmA}\mathcal{E}(\rmI_{t-1}).
\end{equation}
In general, the above equation does not hold
for arbitrary encoder networks, and thus, we can regularize the network with the following regularization loss:
\begin{equation}\label{eq:whippo_reg}
    \mathcal{L} = d(\mathcal{E}(\rmI_t), \bar{\rmA}\mathcal{E}(\rmI_{t-1})),
\end{equation}
for some distance measure $d(\cdot,\cdot)$.

\input{figures/method}

\vspace{1mm}
\subsection{Application to image tokenizers }\label{sec:ae_application}
In this section, we introduce how the regularization loss of Eq.~(\ref{eq:whippo_reg}) is implemented to train an image tokenizer.
Let the tokenizer have a typical autoencoder structure: $f = \mathcal{D} \circ \mathcal{E}$.
Since the tokenizer should retain capability to compress and reconstruct the image, we apply the regularization with probability $\alpha$, and let $f$ be trained to reconstruct the image otherwise.
An illustration of the overall framework is provided in Fig.~\ref{fig:architecture}.
Given an input RGB image $\mathbf{I} = \mathbf{I}_0 \in \mathbb{R}^{3 \times H_0 \times W_0}$, 
we define a continuous blurring sequence $\{\mathbf{I}_\tau\}_{\tau\in[0, \tau_\textrm{max}]}$, where $\tau_\textrm{max}$ is a hyperparameter that decides the maximum level of blur.
Then, we sample two images at different blur levels $0 \leq \tau_1 < \tau_2 \leq \tau_\textrm{max}$, where $|\tau_1 - \tau_2| = \Delta_\mathbf{I}
$ 
for a step size
$\Delta_\mathbf{I} \in \mathbb{R}_{+}$.
Between the two sampled images, the less blurry image $\rmI_{\tau_1}$ is passed through the encoder $\mathcal{E}$ to produce $\mathbf{z}_1 := \mathcal{E}(\rmI_{\tau_1}) \in \mathbb{R}^{C \times H \times W}$ 
where $C, H$ and $W$ denote the channel, height, and width dimensions of the encoded feature map $\mathcal{E}(\rmI_{\tau_1})$, respectively.
Next, we apply the update rule, \ie, multiply the matrix $\bar{\rmA} \in \mathbb{R}^{C \times C}$, to the channel dimension of $\mathbf{z}_1$, which we denote as $\hat{\mathbf{z}}_2 := \bar{\rmA} \mathbf{z}_1$.\footnote{
For brevity, we abuse the matrix multiplication notation to refer to this operation. Technically correct notation should be mode-$n$ product: $\mathbf{z}_2 = \mathbf{z}_1 \times_1 \bar{\rmA} \in \mathbb{R}^{C \times H \times W}$.
}
Since we want $\hat{\mathbf{z}}_2$ to match ${\mathbf{z}}_2 := \mathcal{E}(\rmI_{\tau_2})$, we encode $\rmI_{\tau_2}$ as well to produce its latent ${\mathbf{z}}_2$.
Then, the regularization loss $\mathcal{L}$ is defined as follows:
\begin{equation}\label{eq:whippo_loss}
    \mathcal{L} = \lambda_\mathbf{z} d_\mathbf{z}(\mathbf{z}_2, \hat{\mathbf{z}}_2) + \lambda_\rmI d_\rmI\bigl(\rmI_{\tau_2}, \mathcal{D}(\hat{\mathbf{z}}_2)\bigr),
\end{equation}
where $d_{\mathbf{z}}(\cdot,\cdot)$ and $d_{\rmI}(\cdot,\cdot)$ are distance measures used in latent and pixel spaces, and $\lambda_\mathbf{z}$ and $\lambda_\rmI$ are their respective weights.
Note that we involve the decoder $\mathcal{D}$ in Eq.~(\ref{eq:whippo_loss}).
Minimizing $d_{\mathbf{z}}(\mathbf{z}_{2}, \hat{\mathbf{z}}_{2})$ alone can lead to learning a trivial encoder $\mathcal{E}$ (\eg, $\mathcal{E}: \mathbf{I} \mapsto \mathbf{0}$) and involving $\mathcal{D}$ is empirically found to effectively mitigate such mode collapse, making it a favorable design in previous work~\cite{EQVAE, skorokhodov2025improving}.
By minimizing $\mathcal{L}$, we enforce the state-space-like update between the latent representations of equi-distanced images sampled from $\{\mathbf{I}_\tau\}_{\tau\in[0,\tau_\textrm{max}]}$, thereby regularizing the latent features $\{\mathcal{E}(\rmI_\tau)\}_{\tau \in [0, \tau_\textrm{max}]}$ to follow the dynamics of $\{\mathbf{c}(\mathbf{I}_{\tau})\}_{\tau\in[0,\tau_\textrm{max}]}$.

%% file: figures/teaser.tex
\begin{figure}[t!]
    \centering
    \begin{subfigure}{0.52\linewidth}
        \centering
        \includegraphics[width=\linewidth]{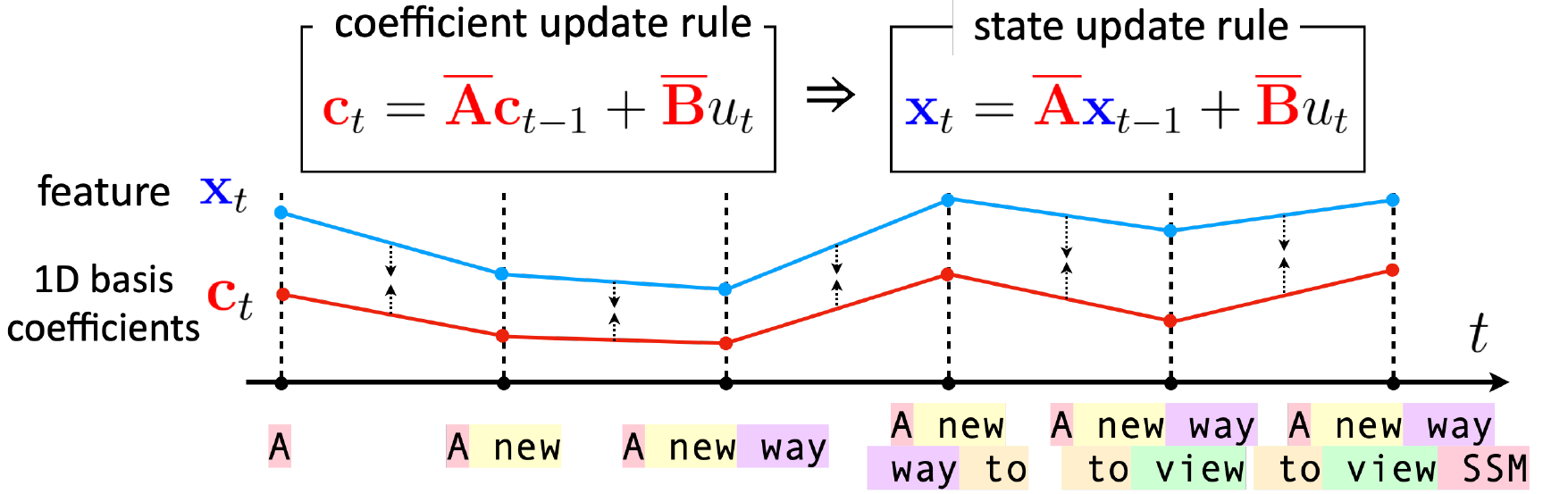}
        \captionsetup{format=hang, indention=0.0mm, singlelinecheck=false}
        \caption{$\mathbf{c}(\cdot)$: Projection onto 1D orthogonal polynomial basis,  \\ $\theta(\cdot)$: Token concatenation}
        \label{fig:hippo}
    \end{subfigure}
    \hfill
    \begin{subfigure}{0.47\linewidth}
        \centering
        \includegraphics[width=\linewidth]{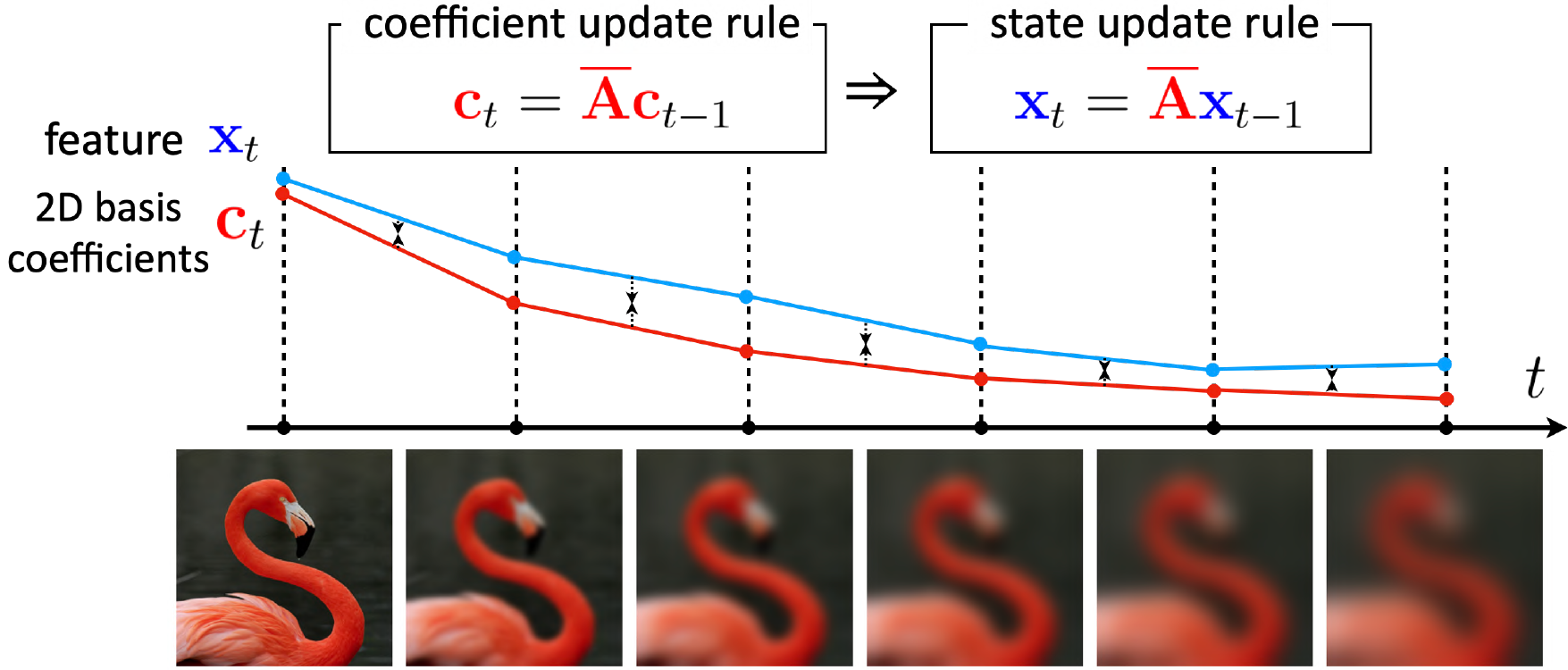}
        \captionsetup{format=hang, indention=0.0mm, singlelinecheck=false}
        \caption{$\mathbf{c}(\cdot)$: Projection onto 2D Fourier basis, \\
        $\theta(\cdot)$: Gaussian blurring}
        \label{fig:whippo}
    \end{subfigure}
    \caption{
    \textbf{Different choice of $\bigl(\mathbf{c}(\cdot), \theta(\cdot) \bigr)$ and the resulting update rules.} 
    Existing SSMs update their hidden state based on Eq.~(\ref{eq:ssm_disc1}), which is derived from the coefficient dynamics based on the choice \textbf{(a)}.
    One can choose a different combination of $\mathbf{c}(\cdot)$ and $\theta(\cdot)$ such as \textbf{(b)}, to derive new dynamics and formulate a new SSM framework.
    }
    \label{fig:teaser}
\end{figure}

%% file: figures/method.tex
\begin{figure}[t!]
    \begin{center}
\includegraphics[width=1.0\linewidth]{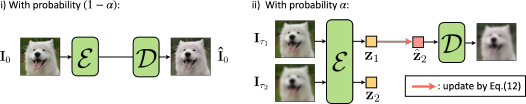}
    \end{center}
\caption{\textbf{State-space regularization applied to an image tokenizer.}
With probability $\alpha$, the update Eq.~(\ref{eq:whippo_update2}) is applied to the latent representation of the input images to produce $\hat{\mathbf{z}}_2$ and $\hat{\mathbf{I}}_{\tau_2}$.
The network is trained to match $(\hat{\mathbf{z}}_2, \hat{\mathbf{I}}_{\tau_2})$ to $(\mathbf{z}_2, \mathbf{I}_{\tau_2})$. 
}
\label{fig:architecture}
\end{figure}

%% file: sections/5_experiments.tex
\vspace{3mm}
\section{Experiments}\label{sec:experiments}

We evaluate the effectiveness of the proposed regularization by training widely used tokenizers, Flux~\cite{Flux} and Cosmos~\cite{CosmosTokenizer}.
Each tokenizer is initialized from their respective pretrained weights and fine-tuned with the reconstruction loss introduced in modern variational autoencoders (VAEs)~\cite{rombach2022high, vqgan} together with the proposed regularization loss Eq.~(\ref{eq:whippo_loss}).
We further train LightningDiT~\cite{vavae, DiT}, a latent diffusion model for image generation, on the resulting latent spaces to evaluate their effectiveness for generation tasks.

\paragraph{Datasets}
We train and evaluate both tokenizers and the diffusion model on ImageNet-1K~\citep{Imagenet}, which contains 1.28M training images and 50K validation images of various animals, objects and scenes.
Following the standard practice, images are resized to 256$^2$ resolution for both training and evaluation.

\paragraph{Implementation details}
The formulation of the proposed regularization varies by the choice of the basis function.
Empirically, we find that using $\mathbf{A}$ derived from the Fourier basis yields the best result.
Unless otherwise specified, all reported results correspond to models using Fourier-based $\mathbf{A}$.
Also, following prior SSM literature, we set the non-zero values of the matrix $\rmA$ and discretization parameter $\Delta$ to be learnable, using a smaller learning rate than other parameters~\citep{gu2022train, gu2022parameterization, gu2021efficiently}.
For tokenizer training, we adopt most hyperparameters from equivariance-regularized autoencoders~\citep{EQVAE, skorokhodov2025improving}, including batch size, learning rate and KL regularization coefficients.
For generation training, we adopt LightningDiT~\citep{vavae}, and train all models for 80K steps, applying classifier-free guidance~\citep{ho2022classifier} at inference with a scale of 10.0.
Further experimental details are provided in Appendix~\ref{appendix:experiment_details}.
Due to spatial constraints, we provide relevant ablation studies in Appendix~\ref{appendix:ablation_studies}.

\paragraph{Evaluation}
We evaluate both reconstruction and generation quality on the ImageNet validation set.
Reconstruction quality is measured using Peak Signal-to-Noise Ratio (PSNR)
and Fréchet Inception Distance (FID), where FID is computed between original images and their corresponding reconstructions (\ie, rFID).
For generation quality, we report generation FID (gFID) and
Inception Score (IS),
widely used 
to assess the perceptual realism of high-resolution generative models.

\subsection{Generation performance comparison}
\input{tables/imagenet_table3}
\input{figures/generation}

We compare reconstruction and generation performance across different methods in Table~\ref{table:imagenet}, including EQVAE~\cite{EQVAE} and FT-SE~\cite{skorokhodov2025improving}, which similarly aim to improve tokenizer's generation-friendliness.
For fair comparisons, we also fine-tune the Flux and Cosmos tokenizers on the ImageNet-1K training set and
use these fine-tuned models to fill the rows for Flux and Cosmos, instead of their original pretrained counterparts.

Table~\ref{table:imagenet} 
compares our method with the previous work in reconstruction and generation performance.
On the Flux tokenizer, our method is on par with or slightly behind other regularizers in reconstruction performance, while yielding larger improvements in generation metrics.
The results on the Cosmos tokenizer show a similar trend, except that all listed regularizers show improved reconstruction quality over the baseline.
Notably, existing regularizers are designed to directly preserve reconstruction-oriented invariances such as scale or rotation consistency, 
whereas our regularization sacrifices the reconstruction training iterations by a factor of $\alpha$.
Despite not explicitly optimizing for reconstruction fidelity, our method achieves competitive reconstruction quality while consistently providing stronger downstream generative performance across most metrics, indicating a favorable tradeoff between reconstruction and generation.
We provide the qualitative examples of generated images in Fig.~\ref{fig:image-generation-compare}.

One noteworthy observation is that our method provides complementary improvements over existing approaches that enhance generation, including ones involving vision foundation models~\citep{vavae, REPA}.
As shown in Table~\ref{table:complementarity}, our method consistently yields additional gains when combined with FT-SE~\citep{skorokhodov2025improving}, VF loss~\citep{vavae}, and REPA~\citep{REPA}.
For the latter two experiments, we use DINOv2~\citep{dinov2} features as the vision foundation model.
Across all three settings, the combined models achieve better generation metrics than either component alone.
This consistent behavior across different generation-improvement strategies suggests that our method captures a complementary source of improvement.

\subsection{Emerging latent space structures from the regularization}\label{sec:latent_structures}
\input{tables/complementarity}
\input{figures/latent_dynamics}

Note that we regularize the latent space by enforcing a basis coefficient's dynamics corresponding to the Gaussian blurring process to the latent feature (Sec.~\ref{sec:gSSM}).
To better understand how the proposed regularization shapes the latent space, we analyze the structures that emerge across latent channels, both qualitatively and quantitatively.
We organize our analysis around three observations.

\vspace{-2mm}
\paragraph{(i) Blurring translates into simpler dynamics in the latent space.}
We randomly sample 100 images from the ImageNet validation set, track their latent representations throughout the image blurring process, and visualize the resulting trajectories using two PCA bases (Fig.~\ref{fig:latent_dynamics}).
Before regularization, the latent trajectories exhibit inconsistency along different blur levels, whereas after applying our regularization, they become notably straighter and more consistent.
This implies that our regularization transforms the latent space such that the progressive frequency decay in pixel space corresponds to simpler and more structured dynamics in the latent space.

\vspace{-2mm}
\paragraph{(ii) Each latent channel encodes what basis coefficients would encode.}
Since each latent channel is enforced to capture a specific frequency band, we examine this behavior by masking selected channels and decoding remaining latent channels back to pixel space.
As we progressively unmask channels from low to high frequencies, the reconstruction correspondingly accumulates coarse-to-fine details, effectively stacking low-to-high frequency components (Left, Fig.~\ref{fig:channel_unveiling}).
We quantitatively examine this behavior by amplifying channels that are grouped by the frequency assigned to each under our derivation (see Appendix~\ref{appendix:channel_assignment} for the grouping details).
Specifically, we divide channels into five groups, $\{g_i\}_{i=1}^{5}$, where a larger $i$ denotes a group assigned to a higher frequency, amplify each group individually (\ie, $\mathbf{z}[c] \mapsto 2(\mathbf{z}[c] - \mu(\mathbf{z}[c]) + \mu(\mathbf{z}[c])$), and identify the frequency band with the largest increase in power (Right, Fig.~\ref{fig:channel_unveiling}).
As shown in Table~\ref{tab:spectral_organization}, our regularization progressively organizes the latent channels according to their intended frequency ordering: as the regularization strength $\alpha$ increases, the frequency bands captured by $g_i$ become increasingly aligned with their prescribed ordering.

\vspace{-1mm}
\paragraph{(iii) Dimensionality of the latent space decreases significantly.}
Fig.~\ref{fig:latent_rank} shows that the effective rank of the latent space is nearly halved after the regularization, while the leading principal components capture a larger portion of the latent variance.
This suggests that the regularized latent space can be well approximated using fewer basis directions, indicating that the regularizer concentrates the latent information into a more compact set of dimensions.

We hypothesize that the enhanced generation-friendliness induced by our regularizer can be attributed to the three aforementioned properties, each supported by related prior studies.
Specifically, (i) well-defined equivariance between the pixel and latent spaces has been shown to improve generation~\citep{skorokhodov2025improving, EQVAE};
(ii) incorporating frequency-aware structure into the latent space potentially benefits the generation process~\citep{dieleman2024spectral, rissanen2022generative};
and (iii) latent distributions that can be represented with fewer Gaussian modes~\citep{chen2025masked} or concentrate their energy into a smaller number of frequency bands~\citep{zhong2026diffusing} have been associated with improved generation.

\input{figures/channel_reveal}
\input{tables/spectral_organization}

%% file: tables/imagenet_table3.tex
\begin{table}[t!]
    \begin{minipage}{\textwidth}
    \caption{\textbf{Reconstruction and generation performance on ImageNet-1K.} Our method improves generation quality across both Flux and Cosmos with minimal impact on reconstruction quality.}
    \vspace{2mm}
    \centering
    \setlength{\tabcolsep}{6.6pt}
    \scalebox{0.85}{
    \begin{tabular}{lcccc}
    \toprule
    \multirow{2}{*}{Tokenizer} & \multicolumn{2}{c}{Reconstruction} & \multicolumn{2}{c}{Generation (-B/2 / -L/2)} \\
    \cmidrule(lr){2-3} \cmidrule(lr){4-5}
    & PSNR $\uparrow$ & rFID $\downarrow$ & gFID $\downarrow$ & IS $\uparrow$ \\
    \midrule
     Flux & \textbf{31.70} & \textbf{0.87} & 16.08 / 5.56 & 89.32 / 184.64 \\
     Flux + EQVAE~\citep{EQVAE} & 31.53 & 0.94 & 13.95 / 5.37 & 97.81 / 195.64 \\
     Flux + FT-SE~\citep{skorokhodov2025improving} & \underline{31.62} & 0.90 & \underline{13.87} / \textbf{5.08} & \underline{99.79} / \underline{201.12} \\
    \rowcolor{gray!20}
    Flux + Ours  & 31.54 & \underline{0.90} & \textbf{13.17} / \underline{5.25} & \textbf{105.99} / \textbf{207.14} \\
    \midrule
    Cosmos & 25.16 & 2.88 & 13.52 / 5.49 & 112.48 / 230.59 \\
    Cosmos + EQVAE~\citep{EQVAE} & \underline{25.20 }& \underline{2.85} & \underline{11.66} / 5.45  & \underline{123.54} / 231.70 \\
    Cosmos + FT-SE~\citep{skorokhodov2025improving} & \textbf{25.37} & 2.87 & 12.01 / \underline{5.36} & 117.97 / \underline{234.18} \\
    \rowcolor{gray!20}
    Cosmos + Ours & 25.18 & \textbf{2.82} & \textbf{11.20} / \textbf{5.23} & \textbf{128.91} / \textbf{237.69} \\
    \bottomrule
    \end{tabular}
    }
    \label{table:imagenet}
    \end{minipage}
\end{table}

%% file: figures/generation.tex
\begin{figure*}[t!]
    \centering
        \includegraphics[width=0.95\linewidth]{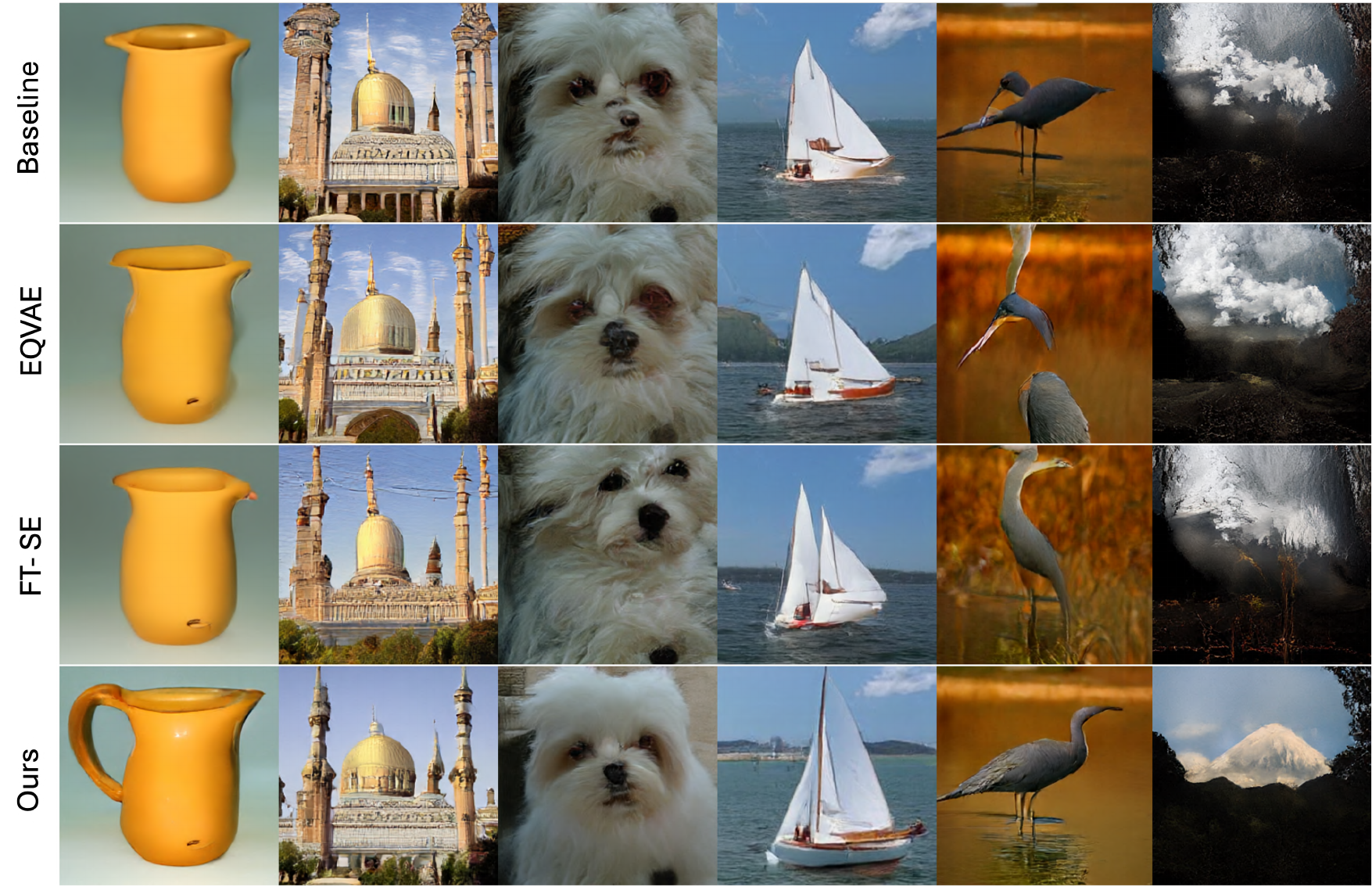}
        \caption{\textbf{Generation results comparison using the Flux tokenizer.} With only marginal loss in reconstruction quality, our method improves the generative performance of the image tokenizer.}
        \label{fig:image-generation-compare}
        \vspace{-2mm}
\end{figure*}

%% file: tables/complementarity.tex
\begin{table}[t!]
    \caption{\textbf{Complementarity with prior generation improvements.}
      Our method complements (a) FT-SE, (b) VF loss, and (c) REPA, yielding additional improvements along orthogonal directions.}
    \label{table:complementarity}
    \centering
    \begin{subtable}[t]{0.34\textwidth}
        \centering
        \caption{FT-SE~\citep{skorokhodov2025improving}}
        \label{table:ft-se_ours_combined}
        \setlength{\tabcolsep}{1.1pt}
        \scalebox{0.70}{
            \begin{tabular}{lcccc}
                \toprule
                 Tokenizer & PSNR $\uparrow$ & {rFID} $\downarrow$ & gFID $\downarrow$ & IS $\uparrow$ \\
                \midrule
                 Flux & \textbf{31.70} & \textbf{0.87} & {16.08} & {89.32} \\
                 \ \ + FT-SE & \underline{31.62} & {0.90} & {13.87} & {99.79} \\
                 \ \ + Ours & {31.54} & \underline{0.90} & \underline{13.17} & \underline{105.99} \\
                 \rowcolor{gray!20}
                 \ \  + FT-SE + Ours & {31.50} & {0.94} & {\textbf{12.46}} & {\textbf{109.58}} \\
                \bottomrule
            \end{tabular}
        }
    \end{subtable}
    \hspace{1.2mm}
    \begin{subtable}[t]{0.34\textwidth}
        \centering
        \caption{VF loss~\citep{vavae}}
        \label{table:vavae_ours_combined}
        \setlength{\tabcolsep}{1.0pt}
        \scalebox{0.70}{
            \begin{tabular}{lcccc}
                \toprule
                 Tokenizer & PSNR $\uparrow$ & {rFID} $\downarrow$ & gFID $\downarrow$ & IS $\uparrow$ \\
                \midrule
                 LDM & \textbf{31.03} & \underline{0.17} & {36.83} & {N/A} \\
                 \ \ + VF loss & \underline{30.71} & \textbf{0.15} & \underline{24.00} & {N/A} \\
                 \ \  + Ours & {30.34} & {0.23} & {26.66} & \underline{82.43} \\
                 \rowcolor{gray!20}
                 \ \  + VF loss + Ours & {30.37} & {0.24} & {\textbf{21.78}} & {\textbf{102.37}} \\
                \bottomrule
            \end{tabular}
        }
    \end{subtable}
    \hfill
    \begin{subtable}[t]{0.27\textwidth}
        \centering
        \caption{REPA~\citep{REPA}}
        \label{table:repa_ours_combined}
        \setlength{\tabcolsep}{1.2pt}
        \scalebox{0.70}{
            \begin{tabular}{lcc}
                \toprule
                 Method & gFID $\downarrow$ & IS $\uparrow$ \\
                \midrule
                 Flux + LightningDiT & {16.08} & {89.32} \\
                 \ \ + REPA  & \underline{9.46} & \underline{133.87} \\
                 \ \ + Ours & {13.17} & {105.99} \\
                 \rowcolor{gray!20}
                 \ \ + REPA + Ours &  {\textbf{8.71}} & {\textbf{142.26}} \\
                \bottomrule
            \end{tabular}
        }
    \end{subtable}
\end{table}

%% file: figures/latent_dynamics.tex
\begin{figure*}[t!]
    \centering
    \includegraphics[width=1.0\linewidth]{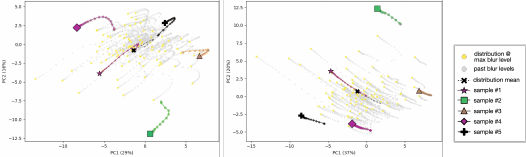}
    \vspace{-3mm}
    \caption{\textbf{Latent dynamics before (left) and after (right) the regularization.}
    Latent trajectories of 100 samples extracted from the ImageNet validation set, projected onto their respective PCA bases.
    Our regularization simplifies the latent dynamics induced by the blurring operation.
    }
    \label{fig:latent_dynamics}
    \vspace{-2mm}
\end{figure*}

% \begin{figure*}[t!]
%     \centering
%     \includegraphics[width=1.0\linewidth]{figures/images/latent_dynamics6.pdf}
%     \caption{\textbf{Latent channel dynamics resmebling the coefficient dynamics.} The top row shows the blurring sequence $\{\mathbf{I}_t\}_{t=0}^{9}$ of an image from the ImageNet validation set, and the next four rows show how the first four channels of the corresponding latent features change as the image blurs.
%     We additionally provide how Fourier coefficients evolve for reference.
%     }
%     \label{fig:latent_dynamics}
% \end{figure*}

%% file: figures/channel_reveal.tex
\begin{figure*}[t!]
    \centering
    \includegraphics[width=1.0\linewidth]{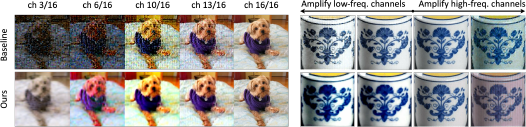}
    \vspace{-3mm}
    \caption{\textbf{Latent channels encoding different frequency bands.} 
    (Left) The regularized tokenizer is able to reconstruct recognizable images solely from using three low-frequency channels.
    (Right) Amplifying each channel and decoding the latent back to pixel space reveals that each channel captures its assigned frequency component as intended.
    }
    \label{fig:channel_unveiling}
    \vspace{-2mm}
\end{figure*}

%% file: tables/spectral_organization.tex
\begin{table}[t!]
    \begin{minipage}[t]{0.31\linewidth}
        \centering
        \captionof{table}{\textbf{Frequency bands captured by each channel group with respect to $\alpha$.}
        Larger regularization strength $\alpha$ encourages channels to encode in an intended spectral order, leading to more spectrally organized representations.}\label{tab:spectral_organization}
        \vspace{2mm}
        \scalebox{0.8}{
        \setlength{\tabcolsep}{3.5pt}
        \begin{tabular}{p{0.8cm}p{0.6cm}p{0.6cm}p{0.6cm}p{0.65cm}p{0.6cm}}
        \toprule
         \multirow{2}{*}{$\alpha$} &  \multicolumn{5}{c}{channel groups}\\
         \cmidrule{2-6}
         & \multicolumn{1}{c}{$g_1$} & \multicolumn{1}{c}{$g_2$} & \multicolumn{1}{c}{$g_3$} & \multicolumn{1}{c}{$g_4$} & \multicolumn{1}{c}{$g_5$} \\
        \midrule
         0.0 & \heat{24.1} & \heat{76.9} & \heat{2.2} & \heat{107.1} & \heat{10.9} \\
         0.05 & \heat{1.5} & \heat{73.1} & \heat{47.2} & \heat{81.2} & \heat{81.4} \\
         0.1 & \heat{1.5} & \heat{29.3} & \heat{47.2} & \heat{81.6} & \heat{81.0} \\
         0.25 & \heat{1.5} & \heat{26.8} & \heat{47.0} & \heat{79.8} & \heat{81.2} \\
         0.5 & \heat{1.5} & \heat{26.1} & \heat{46.6} & \heat{82.6} & \heat{81.2} \\
        \bottomrule
        \end{tabular}
        }
        \vspace{-4mm}
    \end{minipage}
    \hfill
    \begin{minipage}[t]{0.66\linewidth}
        \vspace{4pt}
        \centering
        \includegraphics[width=\linewidth]{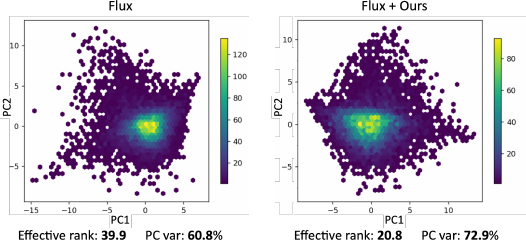}
        \vspace{4pt}
        \captionof{figure}{\textbf{Latent space dimensionality comparison.} 
        Our method reduces the effective rank while increasing the variance captured by the leading principal components of the latent space.}\label{fig:latent_rank}
        \vspace{-3mm}
    \end{minipage}
\end{table}

%% file: sections/6_conclusion.tex
\section{Conclusion}\label{sec:conclusion}
We introduced structured state-space regularization, a framework that extends state-space models (SSMs) to regularize image tokenizers toward generation-friendly representations.
By reframing SSMs as a combination of basis projection and input transformation, we derive a regularization objective that transfers their frequency-aware structure to visual tokenizers.
This formulation bridges state-space modeling and visual representation learning, allowing tokenizers to learn latent representations that effectively capture spectral structures of images.
Experiments show that our method produces latent spaces that better support generative modeling while minimally sacrificing reconstruction quality.
By offering a new perspective on leveraging SSMs for visual representation, this paper paves the way for integrating state-space modeling principles into future generative and representation learning frameworks.

\section*{Acknowledgment}
This work was supported by 
the NRF grant (RS-2026-25618525), % 개인기초핵심전략 - 20%
the IITP grants 
(RS-2022-II220290, % 시각상식 - 10%
RS-2022-II220926, % 자기주도 - 10% 
RS-2024-00509258, RS-2024-00469482, % AI 프론티어 - 10%
RS-2024-00457882, % AI 거점 - 10%
RS-2026-25518317, % 인간인지메모리 - 10%
RS-2019-II191906), % 인공지능대학원
the Advanced GPU Utilization Support Program (02-26-01-0372), and % 고성능 GPU 지원사업 NIPA
the AI Computing Support Project for R\&D (RS-2026-25505492)
funded by Ministry of Science and ICT, Korea.

%% file: sections/10_supplementary.tex
\clearpage
\setcounter{page}{1}
\onecolumn
\begin{center}
    {
    \Large \textbf{
    Structured State-Space Regularization for \\ Generation-Friendly Image Tokenization
    } \\[0.5em]
    }
\end{center}
\appendix
\section{Appendix}
\label{sec:appendix}
We provide supporting details in the Appendix, omitted in the main manuscript due to space constraints.

\input{sections/2_related_work}

\subsection{Combination of $\mathbf{c}(\cdot)$ and $\theta(\cdot)$ that yields tractable update rule}\label{appendix:compression_functions}
In Sec.~\ref{sec:gSSM}, we mention that some choice of basis functions makes it easier to yield a tractable update rule.
Here, the \textit{tractable} update rule refers to having a property where we can directly update the compressed representation to reflect the transformation on the original input \textit{without having to decompress the input}.
Basis projections are favorable in terms of this tractability, since they are in general differentiable, and we can compute the dynamics of the resulting coefficients by differentiating them with respect to $t$.
For example, one can come up with different compressed representations other than basis projection, such as PNG~\cite{boutell1997png} or JPEG~\cite{wallace1991jpeg}.
However, PNG compression is based on entropy-coding, which does not allow direct image-domain linear algebra that corresponds to operations on PNG bitstream.
Thus, PNG necessitates decompression to update the compressed representation, making the update rule intractable.
On the other hand, JPEG is essentially an operation of splitting an image into 8 $\times$ 8 blocks and applying discrete cosine transform~\cite{DCT} to each block.
This indicates that JPEG representation holds frequency components of each block of the image, so the image transformation $\theta: \rmI_{t-1} \mapsto \rmI_t$ must be related to each block's frequency domain.
Thus, a natural choice of $\{\rmI_t \}$ corresponding to JPEG is to apply transformation that involves frequency domain, \eg, gradual blur, on each patch.

As we have seen from the last example, we can conclude that this tractability is determined by both basis projection $\mathbf{c}(\cdot)$ and input transformation $\theta(\cdot)$.
In general, we find that choosing a discontinuous transformation such as downsampling or a stochastic transformation such as noising makes it hard to derive derivatives of $\mathbf{c}(\cdot)$.
Exploring plausible combinations of $\bigl(\mathbf{c}(\cdot), \theta(\cdot)\bigr)$ that result in a nice form of state update rule would be another interesting direction.

\subsection{Derivation of the update rule $\mathbf{c}(\mathbf{I}_{t-1} \mapsto \mathbf{I}_t)$}\label{appendix:update_derivations}
Let the $k$-th element of the vector $\mathbf{c}(\mathbf{I}_\tau) \in \mathbb{R}^N$ be $\mathbf{c}_k(\mathbf{I}_\tau)$, then 
we can derive its derivative as follows:
\begin{align}\label{eq:deriv1}
    \mathbf{c}_k(\mathbf{I}_\tau) &= \langle \mathbf{I}_{\tau}, \phi_k \rangle = \langle G_{\tau} *\mathbf{I}, \phi_k \rangle,\\
    \Rightarrow \ \frac{\mathrm{d}}{\mathrm{d}\tau}\mathbf{c}_k(\mathbf{I}_\tau) &= \frac{\mathrm{d}}{\mathrm{d}\tau}
    \langle G_{\tau} * \mathbf{I}, \phi_k \rangle = \Bigl\langle \frac{\partial}{\partial \tau}(G_{\tau} * \mathbf{I}), \phi_k \Bigr\rangle.\label{eq:heat_eq}
\end{align}
Meanwhile, Gaussian blur kernel $G_{\tau}$ applied to an image $\mathbf{I}$ is known to be the solution to the heat equation~\citep{perona2002scale}, which yields 
\begin{align}
    \Bigl\langle \frac{\partial}{\partial \tau}(G_\tau * \mathbf{I}), \phi_k \Bigr\rangle = \Bigl\langle \frac{1}{2}\nabla^2(G_\tau * \mathbf{I}), \phi_k \Bigr\rangle,\label{eq:heat_eq2}
\end{align}
where $\nabla^2$ is the Laplacian operator.
Since we can express $G_\tau * \mathbf{I}$ by the weighted sum using bases $\{\phi_n\}_{n=1}^N$ and coefficients $\bigl\{\mathbf{c}_n(G_\tau * \mathbf{I})\bigr\}_{n=1}^{N} = \bigl\{\mathbf{c}_n(\mathbf{I}_\tau)\bigr\}_{n=1}^{N}$,
\begin{align}
    \Bigl\langle \frac{1}{2}\nabla^2(G_\tau * \mathbf{I}), \phi_k \Bigr\rangle &= \Bigl\langle \frac{1}{2} \sum_{n=1}^{N}\mathbf{c}_n(\mathbf{I}_\tau)\nabla^2\phi_n, \phi_k \Bigr\rangle \\
    &= \frac{1}{2} \sum_{n=1}^{N}\mathbf{c}_n(\mathbf{I}_\tau) \bigl\langle \nabla^2\phi_n, \phi_k \bigr\rangle.
\end{align}
Thus, gathering all together yields:
\begin{align}
    \frac{\mathrm{d}}{\mathrm{d}\tau}\mathbf{c}_k(\mathbf{I}_\tau) &= \frac{1}{2} \sum_{n=1}^{N}\mathbf{c}_n(\mathbf{I}_\tau) \Bigl\langle \nabla^2 \phi_n, \phi_k \Bigr\rangle.
\end{align}
Aggregating $\mathbf{c}_k$ for all $k \in \{1,2, \ldots , N\}$ gives:
\begin{align}
    \frac{\mathrm{d}}{\mathrm{d}\tau}\mathbf{c}(\mathbf{I}_\tau) 
     &= \frac{1}{2}
    \begin{bmatrix}
        \langle  \nabla^2\phi_1, \phi_1 \rangle & \cdots & \langle \nabla^2\phi_{N}, \phi_{1} \rangle \\
        \langle  \nabla^2\phi_1, \phi_2 \rangle & \cdots & \langle \nabla^2\phi_{N}, \phi_{2} \rangle \\
        \vdots & \ddots & \vdots \\
        \langle  \nabla^2\phi_{1}, \phi_{N} \rangle & \cdots & \langle  \nabla^2\phi_{N}, \phi_{N} \rangle
    \end{bmatrix} \mathbf{c}(\mathbf{I}_\tau) \\
    &:= \mathbf{A} \mathbf{c}(\mathbf{I}_\tau). \label{eq:whippo_dynamics_v2}
\end{align}
Based on the derivative of $\mathbf{c}(\mathbf{I}_\tau)$, we can apply discretization to define the update rule $\mathbf{c}(\mathbf{I}_{t-1}) \mapsto \mathbf{c}(\mathbf{I}_t)$.
We adopt either Euler or ZOH discretization method depending on the choice of the basis functions $\{\phi_n\}_{n=1}^{N}$, following the common practice in modern SSMs~\citep{gu2024mamba, smithsimplified, gupta2022diagonal, gu2022parameterization}.
For a discretization step $\Delta$, we get
\begin{align}
    \textrm{(Euler)}\quad\mathbf{c}(\mathbf{I}_{t}) &= (I + \mathbf{A}\Delta)\mathbf{c}(\mathbf{I}_{t-1}), \label{eq:euler_disc}\\
    \textrm{(ZOH)}\quad\mathbf{c}(\mathbf{I}_{t}) &= e^{\mathbf{A}\Delta}\mathbf{c}(\mathbf{I}_{t-1}), \label{eq:zoh_disc}
\end{align}
which describe the update rules that corresponds to the choice of 2D basis projection $\mathbf{c}(\cdot)$ and Gaussian blurring process $\theta(\cdot)$.
Following the convention of Eq.~(\ref{eq:euler_disc_example}), we denote $\bar{\mathbf{A}} := \delta(\Delta, \mathbf{A})$ for a discretization method $\delta$, and write the update rule after discretization as: 
\begin{equation}\label{eq:whippo_update_appendix}
    \mathbf{c}(\mathbf{I}_t) = \bar{\mathbf{A}}\mathbf{c}(\mathbf{I}_{t-1}).
\end{equation}

We derive the matrix $\mathbf{A}$ for Fourier bases as well as for several families of orthogonal polynomials including Chebyshev, Legendre, and Hermite.
The detailed derivation and results for each basis are provided in Appendix~\ref{appendix:A_derivations}.

\subsection{Derivation based on Gaussian deblurring process}\label{appendix:deblurring}
One can derive the update rule based on Gaussian deblurring process by letting the input image $\mathbf{I} := \mathbf{I}_T$.
For $\tau \in (0, T]$, we adjust $\mathbf{I}_\tau$ of Eq.~(\ref{eq:gaussian_convolution}) as follows:
\begin{align}
    \mathbf{I}_{\tau} = G_{T-\tau} * \mathbf{I}.
\end{align}
Then, the derivative of $\mathbf{c}_k(\mathbf{I}_\tau)$ becomes:
\begin{gather}\label{eq:deblur_deriv1}
    \mathbf{c}_k(\mathbf{I}_\tau) = \langle \mathbf{I}_{\tau}, \phi_k \rangle = \langle G_{T-\tau} *\mathbf{I}, \phi_k \rangle \\
    \Rightarrow \ \frac{\mathrm{d}}{\mathrm{d}\tau}\mathbf{c}_k(\mathbf{I}_\tau) = \frac{\mathrm{d}}{\mathrm{d}\tau}
    \langle G_{T-\tau} * \mathbf{I}, \phi_k \rangle = \Bigl\langle \frac{\partial}{\partial \tau}(G_{T-\tau} * \mathbf{I}), \phi_k \Bigr\rangle
    = - \Bigl\langle \frac{\partial}{\partial s}(G_{s} * \mathbf{I}), \phi_k \Bigr\rangle.\label{eq:deblur_heat_eq}
\end{gather}
Applying the same logic from Sec~\ref{sec:whippo} yields
\begin{align}
    \Bigl\langle \frac{\partial}{\partial s}(G_s * \mathbf{I}), \phi_k \Bigr\rangle = \Bigl\langle \frac{1}{2}\nabla^2(G_s * \mathbf{I}), \phi_k \Bigr\rangle,
\end{align}
where $\nabla^2$ is the Laplacian operator.
Substituting $G_s * \mathbf{I}$ with the linear combination of bases $\{\phi_n\}_{n=1}^N$ using coefficients $\mathbf{c}_n(G_s * \mathbf{I}) = \mathbf{c}_n(G_{T-\tau} * \mathbf{I}) =\mathbf{c}_n(\mathbf{I}_\tau)$ gives:
\begin{align}
    \Bigl\langle \frac{1}{2}\nabla^2(G_s * \mathbf{I}), \phi_k \Bigr\rangle = \Bigl\langle \frac{1}{2} \sum_{n=1}^{N}\mathbf{c}_n(\mathbf{I}_\tau)\nabla^2\phi_n, \phi_k \Bigr\rangle
    = \frac{1}{2} \sum_{n=1}^{N}\mathbf{c}_n(\mathbf{I}_\tau) \bigl\langle \nabla^2\phi_n, \phi_k \bigr\rangle.
\end{align}
Thus, gathering all together yields:
\begin{align}
    \frac{\mathrm{d}}{\mathrm{d}\tau}\mathbf{c}_k(\mathbf{I}_\tau) &= -\frac{1}{2} \sum_{n=1}^{N}\mathbf{c}_n(\mathbf{I}_\tau) \Bigl\langle \nabla^2 \phi_n, \phi_k \Bigr\rangle \\
    \Rightarrow \frac{\mathrm{d}}{\mathrm{d}\tau}\mathbf{c}(\mathbf{I}_\tau) 
     &= -\frac{1}{2}
    \begin{bmatrix}
        \langle  \nabla^2\phi_1, \phi_1 \rangle & \langle  \nabla^2\phi_2, \phi_1 \rangle & \cdots & \langle \nabla^2\phi_{N}, \phi_{1} \rangle \\
        \langle  \nabla^2\phi_1, \phi_2 \rangle & \langle  \nabla^2\phi_2, \phi_2 \rangle & \cdots & \langle \nabla^2\phi_{N}, \phi_{2} \rangle \\
        \vdots & \vdots & \ddots & \vdots \\
        \langle  \nabla^2\phi_{1}, \phi_{N} \rangle & \langle  \nabla^2\phi_{2}, \phi_{N} \rangle & \cdots & \langle  \nabla^2\phi_{N}, \phi_{N} \rangle
    \end{bmatrix} \mathbf{c}(\mathbf{I}_\tau) \\
    &:= -\mathbf{A} \mathbf{c}(\mathbf{I}_\tau),
\end{align}
which is a sign-reversed version of the original dynamics introduced in Eq.~(\ref{eq:whippo_dynamics}).
For a discretization step $\Delta$, we get
\begin{align}
    \textrm{(Euler)}\quad\mathbf{c}(\mathbf{I}_{t}) &= (I - \mathbf{A}\Delta)\mathbf{c}(\mathbf{I}_{t-1}), \\
    \textrm{(ZOH)}\quad\mathbf{c}(\mathbf{I}_{t}) &= e^{-\mathbf{A}\Delta}\mathbf{c}(\mathbf{I}_{t-1}),
\end{align}
Adopting the discretization function $\delta(\cdot, \cdot)$ from Sec.~\ref{sec:gSSM}, we obtain the update of the deblurring process:
\begin{align}
    \mathbf{c}(\mathbf{I}_{t}) &= \delta(-\Delta, \mathbf{A})\mathbf{c}(\mathbf{I}_{t-1}) \\
    &:= \overline{\mathbf{A}}_{\textrm{deblur}}\mathbf{c}(\mathbf{I}_{t-1}).\label{eq:deblur_update}
\end{align}

\subsubsection{What makes it inadequate to apply the deblurring process}\label{appendix:deblurring_problems}
Since the original dynamics based on the Gaussian blurring process (Eq.~(\ref{eq:whippo_update})) does not differ much from the update introduced in Eq.~(\ref{eq:deblur_update}), the deblurring process may seem straightforwardly applicable to the current framework.
However, two critical issues arise when employing the Gaussian deblurring process instead of the blurring process.
First, Gaussian deblurring is fundamentally an \textit{ill-posed} problem: although we obtain the derivative of coefficients $\mathbf{c}(\mathbf{I}_\tau)$ with respect to $\tau$, predicting $\mathbf{c}(\mathbf{I}_{\tau_0+\Delta})$ from $\mathbf{c}(\mathbf{I}_{\tau_0})$ for a timestep $\tau_0$ and a step size $\Delta$ is not a well-defined problem unless the input image has a special constraint, such as being exactly identical to a 2D polynomial of degree $N$~\cite{hummel1985gaussian}.
On the other hand, the blurring process is a \textit{deterministic} process, where having the coefficients of the current timestep $\tau_0$ already suffices to provide a fixed trajectory of $\mathbf{c}(\mathbf{I}_{\tau})\vert_{\tau > \tau_0}$.
This difference ensures the update of the blurring process Eq.~(\ref{eq:whippo_update}) can accurately predict the next coefficients given a sufficiently small step size $\Delta$, whereas Eq.~(\ref{eq:deblur_update}) cannot.
One may introduce an additional information gap $u_t = \mathbf{I}_t - \mathbf{I}_{t-1}$ to modify the deblurring process into a deterministic process, making the formulation more coherent with the original SSM equation Eq.~(\ref{eq:ssm_disc1}).
Yet, under the same assumption of occupying a 2D orthogonal basis projection $\mathbf{c}(\cdot)$, the update becomes very trivial due to the linearity of basis functions:
\begin{equation}
    \mathbf{c}(\mathbf{I}_{t}) = \mathbf{c}(\mathbf{I}_{t-1} + u_t) = \mathbf{c}(\mathbf{I}_{t-1}) + \mathbf{c}(u_t),
\end{equation}
which turns into a feature dynamics expressed as $f(\mathbf{I}_t) = f(\mathbf{I}_{t-1}) + f(u_t)$ for a feature extractor $f$.
Here, $f$ inherits one of the fundamental properties of the basis projection $\mathbf{c}(\cdot)$, linearity, but it is not a unique property of basis projection and $f$ hardly achieves our goal of obtaining frequency awareness and compactness via mimicking the behavior of $\mathbf{c}(\cdot)$.
Choosing a different pair of $\bigl(\mathbf{c}(\cdot), \phi(\cdot)\bigr)$ may lead to a different conclusion and yield a formulation that is more coherent to the original 1D SSMs, but for now we leave it for future work.
Second, when applied to a tokenizer's encoder $\mathcal{E}$, Eq.~(\ref{eq:deblur_update}) hinders encoder's ablity to capture details.
Note that the amount of information in the blurred image's feature $f(\mathbf{I}_\mathrm{blurred})$ cannot be amplified by simply multiplying with $\bar{\mathbf{A}}$ to match that of  $f(\mathbf{I}_\mathrm{sharp})$.
Since the regularization loss based on Eq.~(\ref{eq:deblur_update}) forces $f(\mathbf{I}_\mathrm{sharp})$ to match $\bar{\mathbf{A}}f(\mathbf{I}_\mathrm{blurred})$ during training, $f$ ends up encoding only as much information as the blurred input, even when given a sharp image.

\subsection{Relation to the heat equation}\label{appendix:heat_equation}
The heat equation is a partial differential equation that illustrates thermodynamics.
Formally, let $T(x,y,t): [0,W] \times [0,H] \times [0,\infty) \rightarrow \mathbb{R}$ be a surface function that indicates a temperature at the point $(x,y)$ at time $t$. Then, the heat equation is defined as follows:
\begin{equation}
    \frac{\partial T}{\partial t}= \alpha \Bigl(\frac{\partial^2 T}{\partial x^2} + \frac{\partial^2 T}{\partial y^2}\Bigr) := \alpha \nabla^2 T,
\end{equation}
where $\alpha$ is a positive constant that decides the speed of heat diffusion.
Let $T(x,y,0)$ be the image $\mathbf{I}(x,y)$ and $\left.\frac{\partial T}{\partial x}\right|_{x\in \{0, W\}} = \left.\frac{\partial T}{\partial y}\right|_{y\in \{0, H\}} = 0$,
then its known solution $u(x,y,t)$ is:
\begin{equation}
    u(x,y,t) = G_t * \mathbf{I}, 
\end{equation}
where $G_{\sigma^2}(x,y) = \frac{1}{4\alpha\pi \sigma^2}\exp(-\frac{x^2 + y^2}{4\alpha\sigma^2})$.
Intuitively, if the pixel intensity at coordinate $(x,y)$ is viewed as temperature, then the diffusion of heat over time corresponds precisely to the Gaussian-blurred image.
Hence, the rightmost term $G_{s} * \mathbf{I}$ of Eq.~(\ref{eq:heat_eq}) satisfies $\frac{\partial T}{\partial t}= \frac{1}{2} \Bigl(\frac{\partial^2 T}{\partial x^2} + \frac{\partial^2 T}{\partial y^2}\Bigr) := \frac{1}{2} \nabla^2 T$, which explains the equality in Eq.~(\ref{eq:heat_eq2}).

\subsection{Derivation of \textbf{A} matrices}\label{appendix:A_derivations}
Let us restate the state transition matrix $\mathbf{A}$ for reference:
\begin{equation}
    \mathbf{A} = \frac{1}{2}
    \begin{bmatrix}
        \langle \nabla^2\phi_1, \phi_1 \rangle & \langle \nabla^2\phi_2, \phi_1 \rangle & \cdots & \langle \nabla^2\phi_{N}, \phi_{1} \rangle \\
        \langle \nabla^2\phi_1, \phi_2 \rangle & \langle \nabla^2\phi_2, \phi_2 \rangle & \cdots & \langle \nabla^2\phi_{N}, \phi_{2} \rangle \\
        \vdots & \vdots & \ddots & \vdots \\
        \langle \nabla^2\phi_{1}, \phi_{N} \rangle & \langle \nabla^2\phi_{2}, \phi_{N} \rangle & \cdots & \langle \nabla^2\phi_{N}, \phi_{N} \rangle
    \end{bmatrix}.
\end{equation}
Here, $\{\phi_n\}_{n=1}^{N} = \{\phi_{w,h}\}_{(w,h) \in \{0,1,\ldots,W-1\} \times \{0,1,\ldots,H-1\}}$ is a set of 2D basis functions.
For each index $n$ in $\{1,2, \ldots N\}$, we specifically define the mapping function $\iota$ between two index sets $\{1,2, \ldots, N\}$ and $\{(0, 0), \ldots, (W-1, H-1)\}$ as follows:
\begin{equation}\label{eq:index_mapping}
    \iota: n \mapsto \bigl((n-1)~\bmod W, \lfloor \frac{n-1}{W}\rfloor  \bigr).
\end{equation}
As we obtain a different form of $\mathbf{A}$ depending on the choice of the basis function $\phi$, we introduce derivations for different basis choices: Fourier, Chebyshev polynomial, Legendre polynomial, and Hermite polynomial.
Since deriving $\mathbf{A}$ is equivalent to computing $\langle \phi_m, \nabla^2\phi_n \rangle$ (or equivalently $\langle \phi_{w_1, h_1}, \nabla^2\phi_{w_2, h_2} \rangle$), we focus on deriving the inner product between a basis function and the second derivative of another basis function.
We additionally display each matrix in Fig.~\ref{fig:A_matrices}.

\input{figures/A_matrices}

\subsubsection{Fourier \textbf{A}}\label{appendix:fourier_A}
Note that a 2D Fourier basis defined on $[0, W] \times [0, H]$ is defined as follows:
\begin{equation}
    \phi_{w,h}(x,y) = \exp\Bigl(2\pi i \bigl(\frac{wx}{W} + \frac{hy}{H} \bigr)\Bigr).
\end{equation}
We obtain its derivative and second derivative with respect to $x$ as:
\begin{align}
    \frac{\partial\phi_{w,h}(x,y)}{\partial x} &= \frac{2\pi iw}{W} \exp\Bigl(2\pi i \bigl(\frac{wx}{W} + \frac{hy}{H} \bigr)\Bigr), \\
    \frac{\partial^2\phi_{w,h}(x,y)}{\partial x^2} &= -\frac{4\pi^2w^2}{W^2} \exp\Bigl(2\pi i \bigl(\frac{wx}{W} + \frac{hy}{H} \bigr)\Bigr) \notag \\
    &= -\frac{4\pi^2w^2}{W^2}\phi_{w,h}(x,y).
\end{align}
Therefore, $\nabla^2 \phi_{w,h}(x,y)$ becomes:
\begin{align}
    \nabla^2\phi_{w,h}(x,y) &= \frac{\partial^2 \phi_{w,y}(x,y)}{\partial x^2} + \frac{\partial^2 \phi_{w,y}(x,y)}{\partial y^2} \notag \\
    &= -4\pi^2\bigl( \frac{w^2}{W^2} + \frac{h^2}{H^2} \bigr)\phi_{w,h}(x,y).
\end{align}
Note that Fourier bases are orthogonal bases; \ie, $\langle \phi_{w_1,h_1}, \phi_{w_2, h_2} \rangle = 0$ if $ (w_1, h_1) \neq (w_2, h_2)$.
Hence, the matrix element $\langle \phi_{w_1,h_1}, \nabla^2\phi_{w_2, h_2} \rangle$ becomes:
\begin{equation}
    \langle \phi_{w_1,h_1}, \nabla^2\phi_{w_2,h_2} \rangle \\
    =
    \left\{
        \begin{array}{ll}
        0 & (w_1, h_1) \neq (w_2, h_2), \\
        -4WH\pi^2 \bigl( \frac{w_1^2}{W^2} + \frac{h_1^2}{H^2} \bigr) & \text{otherwise}.
        \end{array}
    \right.
    \end{equation}
which makes the matrix $\mathbf{A}$ diagonal.

\subsubsection{Chebyshev \textbf{A}}\label{appendix:chebyshev_A}
A 2D Chebyshev polynomial basis defined on $[0, W] \times [0, H]$ is defined as follows:
\begin{equation}
    \phi_{w,h}(x,y) = \cos\Bigl(w\cos^{-1}\bigl(\frac{2x}{W}-1\bigr)\Bigr) \cdot \cos\Bigl(h\cos^{-1}\bigl(\frac{2y}{H}-1\bigr)\Bigr).
\end{equation}
Let us denote $\phi_{w}(u)$ and $\phi_{h}(v)$ as follows:
\begin{align}
    \phi_{w}(u) &= \cos\bigl(w\cos^{-1}(u)\bigr) \hspace{1em} \forall u \in [-1, 1], \\
    \phi_{h}(v) &= \cos\bigl(h\cos^{-1}(v)\bigr) \hspace{1em} \forall v \in [-1, 1],
\end{align}
then $\phi_{w,h}(x,y) = \phi_w(\frac{2x}{W}-1) \cdot \phi_h(\frac{2y}{H}-1)$.
We obtain its derivative and second derivative with respect to $x$ as:
\begin{align}
    \frac{\partial{\phi_{w,h}(x,y)}}{\partial x} &= \frac{2}{W}\phi_w'\bigl(\frac{2x}{W}-1\bigr)\cdot \phi_h\bigl(\frac{2y}{H}-1\bigr), \\
\frac{\partial^2{\phi_{w,h}(x,y)}}{\partial x^2} &= \frac{4}{W^2}\phi_w''\bigl(\frac{2x}{W}-1\bigr)\cdot \phi_h\bigl(\frac{2y}{H}-1\bigr).
\end{align}
Therefore, $\nabla^2 \phi_{w,h}(x,y)$ becomes:
\begin{equation}
    \nabla^2{\phi_{w,h}(x,y)} = \frac{4}{W^2}\phi_w''\biggl(\frac{2x}{W}-1\biggr)\cdot \phi_h\biggl(\frac{2y}{H}-1\biggr)
    + \frac{4}{H^2}\phi_w\biggl(\frac{2x}{W}-1\biggr)\cdot \phi_h''\biggl(\frac{2y}{H}-1\biggr).
\end{equation}
Reparameterize $(u, v) = \bigl(\frac{2x}{W} - 1, \frac{2y}{H} - 1 \bigr)$, and let the weight function of the Chebyshev polynomial $\omega(u, v) = \frac{1}{\sqrt{1-u^2}}\cdot\frac{1}{\sqrt{1-v^2}} := \omega(u)\cdot \omega(v)$.
Then, $\langle \phi_{w_1,h_1}, \nabla^2\phi_{w_2, h_2} \rangle_\omega$ becomes:
\begin{align}\label{eq:cheby}
    & \langle \phi_{w_1,h_1}, \nabla^2\phi_{w_2,h_2} \rangle_\omega \nonumber \\
    &=
    \frac{W}{2} \cdot \frac{H}{2} \iint\limits_{[-1,1]^2}
    \phi_{w_1}(u)\phi_{h_1}(v)
    \Biggl(
        \frac{4}{W^2} \phi_{w_2}''(u) \phi_{h_2}(v) + 
        \frac{4}{H^2} \phi_{h_2}''(v) \phi_{w_2}(u)
    \Biggr)
    \omega(u,v) \, \mathrm{d}v\mathrm{d}u \\
    &=
    \frac{H}{W} \iint\limits_{[-1,1]^2}
    \phi_{w_1}(u)\phi_{w_2}''(u)
    \phi_{h_1}(v)\phi_{h_2}(v)
    \omega(u,v) \, \mathrm{d}v\mathrm{d}u \nonumber \\
    & \qquad \qquad \qquad \qquad \qquad +
    \frac{W}{H} \iint\limits_{[-1,1]^2}
    \phi_{w_1}(u)\phi_{w_2}(u)
    \phi_{h_1}(v)\phi_{h_2}''(v)
    \omega(u,v) \, \mathrm{d}v\mathrm{d}u \nonumber \\
    &=
    \frac{H}{W}
        \int_{-1}^{1} \phi_{w_1}(u)\phi_{w_2}''(u)\omega(u) \, \mathrm{d}u
        \cdot
        \int_{-1}^{1} \phi_{h_1}(v)\phi_{h_2}(v)\omega(v) \, \mathrm{d}v \nonumber \\
    & \qquad \qquad \qquad \qquad \qquad +
    \frac{W}{H}
        \int_{-1}^{1} \phi_{w_1}(u)\phi_{w_2}(u)\omega(u) \, \mathrm{d}u
    \cdot
        \int_{-1}^{1} \phi_{h_1}(v)\phi_{h_2}''(v)\omega(v) \, \mathrm{d}v \\
    &=
    \frac{H}{W}
    \left\langle \phi_{w_1}, \phi_{w_2}'' \right\rangle_{\omega}
    \cdot
    \left\langle \phi_{h_1}, \phi_{h_2} \right\rangle_{\omega}
    +
    \frac{W}{H}
    \left\langle \phi_{w_1}, \phi_{w_2} \right\rangle_{\omega}
    \cdot
    \left\langle \phi_{h_1}, \phi_{h_2}'' \right\rangle_{\omega}.
\end{align}
From Shen~\cite[Eq.~(2.9)]{shen1995efficient}, we have
\begin{equation}
    \phi_k''(x) 
    = \sum_{\substack{0 \le n \le k-2 \\ n+k \,\text{even}}} 
      \frac{k(k^2 - n^2)}{c_n} \, \phi_n(x),
    \qquad
    \hspace{-1em} c_n = \begin{cases}
        2 & n=0, \\
        1 & n>0.
    \end{cases}
\end{equation}
Hence,
\begin{equation}
    \langle \phi_j, \phi_k'' \rangle_\omega
    = \Big\langle \phi_j, 
      \sum_{\substack{0 \le n \le k-2 \\ n+k \,\text{even}}} 
      \frac{k(k^2-n^2)}{c_n} \, \phi_n
      \Big\rangle_\omega .
\end{equation}
Also we know by the property of the Chebyshev polynomial that:
\begin{equation}
  \langle \phi_{j}, \phi_{k}\rangle_{\omega} =
  \begin{cases}
    \pi & j = k = 0,\\
    \pi/2 & j = k \neq 0,\\
    0 & j \neq k.
  \end{cases} \label{eq:cheby1}
\end{equation}
Thus, we obtain
\begin{equation}
    \langle \phi_j, \phi_k'' \rangle_\omega = 
    \begin{cases}
        0 & j > k-2, \\[6pt]
        0 & j \not\equiv k \pmod{2}, \\[6pt]
        \dfrac{\pi}{2}\, k (k^2 - j^2) &
        j \equiv k \pmod{2}, \ j \le k-2 .
    \end{cases}
    \label{eq:cheby2}
\end{equation}
By substituting Eq.~(\ref{eq:cheby1}) and Eq.~(\ref{eq:cheby2}) into Eq.~(\ref{eq:cheby}), 
we obtain $\langle \phi_{w_1,h_1}, \nabla^2 \phi_{w_2,h_2} \rangle_\omega$, 
which allows us to construct the matrix $\mathbf{A}$ associated with the Chebyshev polynomials.

\subsubsection{Legendre \textbf{A}}\label{appendix:legendre_A}
Analogous to the derivation of the Chebyshev $\mathbf{A}$, we can also express Legendre basis $\phi_{w,h}$ defined on $[0, W] \times [0, H]$ as  $\phi_{w,h}(x,y) = \phi_w(\frac{2x}{W}-1) \cdot \phi_h(\frac{2y}{H}-1)$.
Then, we can apply the similar logic of Eq.~(\ref{eq:cheby}):
\begin{align}
    \langle \phi_{w_1,h_1}, \nabla^2\phi_{w_2, h_2} \rangle_\omega &= \nonumber \\ \frac{H}{W}\Bigl(\langle \phi_{w_1}, \phi_{w_2}'' &\rangle_{\omega} \cdot \langle \phi_{h_1}, \phi_{h_2} \rangle_{\omega} \Bigr) + \frac{W}{H} \Bigl(\langle \phi_{w_1}, \phi_{w_2} \rangle_{\omega} \cdot  \langle \phi_{h_1}, \phi_{h_2}'' \rangle_{\omega}\Bigr), \label{eq:legendre}
\end{align}
where $\phi_w(u)$ and $\phi_h(v)$ are 
\begin{align}
\phi_w(u) &= \frac{1}{2^ww!}\frac{\mathrm{d}^w}{\mathrm{d}u^w}(u^2-1)^w ~\forall u \in [-1, 1], \\
\phi_h(v) &= \frac{1}{2^hh!}\frac{\mathrm{d}^h}{\mathrm{d}v^h}(v^2-1)^h ~\forall v \in [-1, 1],
\end{align}
respectively.
Note that the weight function of the Legendre polynomial $\omega(u,v) = 1$.
Thus, calculating $\langle \phi_{j}, \phi_{k}\rangle$ and $\langle \phi_{j}, \phi''_{k}\rangle$ suffices to derive Legendre $\mathbf{A}$.
From Shen~\cite[Eq.~(2.6)]{shen1994efficient}, we have
\begin{equation}
    \phi_k''(x) 
    = \sum_{\substack{0 \le n \le k-2 \\ n+k \,\text{even}}} 
      (n+\tfrac{1}{2}) \bigl[k(k+1) - n(n+1)\bigr] \, \phi_n(x).
\end{equation}
Thus,
\begin{equation}
    \langle \phi_j, \phi_k'' \rangle
    = \left\langle \phi_j, 
      \sum_{\substack{0 \le n \le k-2 \\ n+k \,\text{even}}} 
      (n+\tfrac{1}{2}) \bigl[k(k+1) - n(n+1)\bigr] \phi_n \right\rangle.
\end{equation}
By the orthogonality of the Legendre polynomials,
\begin{equation}
    \langle \phi_j, \phi_k \rangle =
    \begin{cases}
        \dfrac{2}{2k+1} & j = k, \\[6pt]
        0 & j \neq k .
    \end{cases}
    \label{eq:legendre1}
\end{equation}
Therefore,
\begin{equation}
    \langle \phi_j, \phi_k'' \rangle =
    \begin{cases}
        0 & j > k-2, \\[6pt]
        0 & j \not\equiv k \pmod{2}, \\[6pt]
        \bigl[k(k+1) - j(j+1)\bigr] & j \equiv k \pmod{2}, \ j \le k-2.
    \end{cases}
    \label{eq:legendre2}
\end{equation}

\subsubsection{Hermite \textbf{A}}
We adopt the physicist's Hermite polynomial defined on $\mathbb{R}$ as follows:
\begin{equation}
    \phi_{m}^{\mathbb{R}}(x) = (-1)^m e^{x^2} \frac{\mathrm{d}^m}{\mathrm{d}x^w}.
\end{equation}
We define the basis whose domain is restricted to $[0, W]$ as:
\begin{align}
    \phi_{m}(x) := \phi_{m}^{[0, W]}(x) = \phi_{m}^{\mathbb{R}}\Bigl(\frac{2x-W}{\alpha W}\Bigr),
\end{align}
where $\alpha$ is a constant that controls how tightly the Hermite basis is concentrated inside the image.
We set $\alpha = \frac{1}{2}$.
Thus, the 2D Hermite basis defined on $[0, W] \times [0, H]$ is:
\begin{equation}
 \phi_{w,h}(x,y) = \phi_w^{\mathbb{R}}\Bigl(\frac{4x}{W}-2\Bigr) \cdot \phi_h^{\mathbb{R}}\Bigl(\frac{4y}{H}-2\Bigr)
\end{equation}
Reparameterize $(u, v) = \bigl(\frac{4x}{W} - 2, \frac{4y}{H} - 2 \bigr)$, and let the weight function of the Hermite polynomial $\omega(u,v) = \frac{e^{-(u^2+v^2)}}{\sqrt{\pi}}$.
Then, by the similar logic of Eq.~(\ref{eq:cheby}), $\langle \phi_{w_1,h_1}, \nabla^2\phi_{w_2, h_2} \rangle_\omega$ becomes:
\begin{align}\label{eq:hermite}
    \langle \phi_{w_1,h_1}, \nabla^2\phi_{w_2, h_2} \rangle_\omega& \\ = 
    \frac{H}{W}\Bigl(\langle \phi^{\mathbb{R}}_{w_1}, \phi^{\mathbb{R}}_{w_2}{}'' &\rangle_{\omega} \cdot  \langle \phi^{\mathbb{R}}_{h_1}, \phi^{\mathbb{R}}_{h_2} \rangle_{\omega} \Bigr) + \frac{W}{H} \Bigl(\langle \phi^{\mathbb{R}}_{w_1}, \phi^{\mathbb{R}}_{w_2} \rangle_{\omega} \cdot  \langle \phi^{\mathbb{R}}_{h_1}, \phi^{\mathbb{R}}_{h_2}{}'' \rangle_{\omega}\Bigr).
\end{align}
Note the recurrence in Hermite polynomial:
\begin{align}
    \phi^{\mathbb{R}}_m{}'(x) &= 2m\phi^{\mathbb{R}}_{m-1}(x), \\
    \phi^{\mathbb{R}}_m{}''(x) &= 4m(m-1)\phi^{\mathbb{R}}_{m-2}(x).
\end{align}
Also, it is known that 
\begin{equation}\label{eq:hermite1}
    \langle \phi^{\mathbb{R}}_{j}, \phi^{\mathbb{R}}_{k}\rangle_{\omega} =\begin{cases}
        \sqrt{\pi}2^kk! & j = k,\\
        0 & j \neq k.
    \end{cases}
\end{equation}
Hence, $\langle \phi^{\mathbb{R}}_{j}, \phi^{\mathbb{R}}_{k}{}''\rangle_{\omega}$ becomes
\begin{equation}\label{eq:hermite2}
    \langle \phi^{\mathbb{R}}_{j}, \phi^{\mathbb{R}}_{k}{}''\rangle_{\omega} =\begin{cases}
        0 & k < 2, \\
        0 & j \neq k-2, \\
        4k(k-1)\sqrt{\pi}2^{k-2}(k-2)! & j = k-2.
    \end{cases}
\end{equation}
Plugging Eq.~(\ref{eq:hermite1}) and Eq.~(\ref{eq:hermite2}) to Eq.~(\ref{eq:hermite}) completes the construction of matrix $\mathbf{A}$ associated with the Hermite polynomial.

\input{tables/experiment_details1}
\input{tables/experiment_details2}

\subsection{Experiment details}\label{appendix:experiment_details}

We provide details of the tokenizer training experiment in Table~\ref{tab:experiment_details1}.
The batch size for both tokenizers and gradient clipping norm are adopted from FT-SE~\cite{skorokhodov2025improving}, while KL weight is set to $10^{-6}$ following EQVAE~\cite{EQVAE}.
We use a cosine annealing learning rate scheduler combined with linear warmup and a constant learning rate scheduler, which linearly raises the learning rate from initialized warmup learning rate to base learning rate, maintain the base learning rate, and then plateau to the minimum learning rate at the end.
Specifically, the warmup period is set to one-tenth of the total training iterations, while the annealing period spans half of the total iterations.
Following the practice in modern SSMs~\cite{gu2021efficiently, smithsimplified}, we apply smaller learning rates to SSM parameters $\mathbf{A}$ and $\Delta$, specifically, one-tenth of the base learning rate.
Borrowing the perspective discussed in Sec.~\ref{sec:gSSM}, occupying smaller learning rates for these parameters indicates regularizing a network to more strictly follow the coefficient dynamics $\mathbf{c}(\mathbf{I}_{t-1}) \mapsto \mathbf{c}(\mathbf{I}_{t})$, since the regularization loss $\mathcal{L}$ (Eq.~(\ref{eq:whippo_reg})) would be minimized merely by optimizing the network parameters rather than dramatically updating the SSM parameters.
Furthermore, we freeze the first two encoder blocks and the last two decoder blocks of tokenizers, similar to Skorokhodov \etal~\cite{skorokhodov2025improving}.
This strategy preserves the general capabilities of the pretrained weights and mitigates overfitting to the target dataset during fine-tuning.
Moreover, it has been empirically shown to effectively replace the need for discriminator loss~\cite{vqgan}, thereby accelerating training~\cite{skorokhodov2025improving}.

Table~\ref{tab:whippo_implementation_details2} shows the hyperparameter setting we use for the regularization.
The maximum level of blur $\tau_\textrm{max}$ on image is set $10.0$, while the gaussian blurring difference between images $\Delta_\mathbf{I}$ is fixed to 4.0.
To illustrate how the magnitude of $\tau$ influences the image, we show the corresponding blurring sequence in Fig.~\ref{fig:img_blurring_sequence}.
When we sample an image pair of distance $\Delta_{\rmI}$ from the continuous interval $[0, \tau_\textrm{max}]$, we assume a linearly decaying probability density: $\tau_1 \sim p(x) = \frac{-2x}{(\tau_\textrm{max} - \Delta_{\rmI})^2} + \frac{2}{\tau_\textrm{max} - \Delta_{\rmI}}~\forall x \in [0, \tau_\mathrm{max} - \Delta_\rmI]$.
Since the perceptual difference between image pairs becomes more subtle as $\tau_1$ increases, sampling smaller $\tau_1$ values provides stronger learning signals for the regularization.
The SSM parameter $\Delta$, which is used for the discretization introduced in Eq.~(\ref{eq:euler_disc}) and Eq.~(\ref{eq:zoh_disc}), is initialized to 0.1.
As mentioned in Fig.~\ref{fig:A_matrices}, the matrix $\mathbf{A}$ is normalized to have the maximum absolute value of 1 and is used as initialization values.
To maintain the structure of $\mathbf{A}$, only non-zero values of the matrix are updated during the training.
For the distance measure in latent space $d_\mathbf{z}$, we use mean squared error (MSE), while those in image space $d_\rmI$ occupies mean absolute error (MAE) combined with learned perceptual image patch similarity (LPIPS).
We find that the optimal weight of the latent space measure $\Delta_\mathbf{z}$ varies by the type of the tokenizer.
Since the loss between $\mathbf{z}_1$ and $\mathbf{z}_2$ is computed by encoding two different images, we take additional methods to prevent mode collapse, \ie $\mathcal{E}: \rmI \mapsto c$ for a constant $c$. 
Specifically, to avoid $\mathbf{z}_1$ collapsing toward the representation of $\mathbf{z}_2$, we stop gradients when encoding $\rmI_{\tau_1}$ to obtain $\mathbf{z}_1$.
Also, the encoder used to produce $\mathbf{z}_1$ is updated using an exponential moving average (EMA) of the main encoder weights with a decay rate of 0.999, thereby letting $\mathbf{z}_1$ not directly affected by dynamically updated weights that is optimized to minimize $d_\mathbf{z}(\mathbf{z}_1, \mathbf{z}_2).$

\input{figures/mean_centered}

\subsection{Modification for practical implementation}\label{sec:practical_implementation}
We find that $\lVert \bar{\mathbf{A}} \rVert < 1$ for all choices of the basis function, which leads to the latent of the blurrier image having much smaller norm than that of the sharper image.\footnote{Note that the latent of a blurry image $\rmI_N$ is represented as $\mathcal{E}(\rmI_N) = 
% \bar{\mathbf{A}}\mathcal{E}(\rmI_{N-1}) = \cdots =
\bar{\mathbf{A}}^N\mathcal{E}(\rmI_0)$, by unrolling Eq.~(\ref{eq:whippo_update2}).}
Although this indeed reflects the decaying behavior of the basis coefficients under blurring operation, in practice, this can cause latent representations of sharp images to exhibit undesirably high norm, which in turn destabilitize training.
To mitigate this issue, we regularize $\mathcal{C}\bigl(\mathcal{E}(\rmI_\tau)\bigr)$ rather than $\mathcal{E}(\rmI_\tau)$, where $\mathcal{C}$ denotes a spatial mean-centering function defined as follows:
\begin{equation}
    \mathcal{C}: \mathbb{R}^{C \times H \times W} \to \mathbb{R}^{C \times H \times W}, \qquad
    \mathcal{C}(\mathcal{\rmI})_{c,h,w} = \rmI_{c,h,w} - \frac{1}{HW}\sum_{\forall(h', w')}\rmI_{c,h',w'}.
\end{equation}
Intuitively, it drives latent values to approach to their respective spatial mean instead of zero as the input image blurs, and thus, the latent norm can avoid decaying towards zero.

Although the centering may appear to alter the target dynamics, the centered features in fact follow the same update rule with the same $\overline{\mathbf{A}}$.
\begin{prop}
    If $\mathcal{E}(\mathbf{I}_t) = \overline{\mathbf{A}}\mathcal{E}(\mathbf{I}_{t-1})$, then $\mathcal{C}\bigl(\mathcal{E}(\mathbf{I}_t)\bigr) = \overline{\mathbf{A}}\mathcal{C}(\mathcal{E}(\mathbf{I}_{t-1}))$.
\end{prop}
\begin{proof}
    Let the latent $\mathcal{E}(\mathbf{I})$ be denoted as $\mathbf{z} \in \mathbb{R}^{C \times HW}$.
    While the update acts on the left, \ie, $\mathbf{z} \mapsto \overline{\mathbf{A}}\mathbf{z}$, the spatial mean-centering $\mathcal{C}$ acts on the right:
    \begin{equation}
        \mathcal{C}: \mathbf{z} \mapsto \mathbf{z}\mathbf{P}, \qquad \textrm{where} \quad \mathbf{P} = \mathbf{I} - \frac{1}{HW}\mathbf{1}\mathbf{1}^{\top} \in \mathbb{R}^{HW \times HW}.
    \end{equation}
    By associativity, $\mathcal{C}(\overline{\mathbf{A}}\mathbf{z}) = \overline{\mathbf{A}}\mathbf{z}\mathbf{P} = \overline{\mathbf{A}}\mathcal{C}(\mathbf{z})$.
    Hence, if $\mathcal{E}(\mathbf{I}_t) = \overline{\mathbf{A}}\mathcal{E}(\mathbf{I}_{t-1})$, applying $\mathcal{C}$ to both sides yields:
    \begin{equation}
        \mathcal{C}\bigl(\mathcal{E}(\mathbf{I}_t)\bigr) = \mathcal{C}\bigl(\overline{\mathbf{A}}\mathcal{E}(\mathbf{I}_{t-1})\bigr) = \overline{\mathbf{A}}\mathcal{C}(\mathcal{E}(\mathbf{I}_{t-1})),
    \end{equation}
    which implies that the centered dynamics is directly obtained if the latent follows our target dynamics.
\end{proof}
% We discuss the related details and provide illustrations in Sec.~\ref{sec:mean_centering} and Fig.~\ref{fig:mean-centered-whole}.
% \subsection{Effect of regularizing $\mathcal{C}\bigl(\mathcal{E}(\rmI_\tau)\bigr)$ instead of $\mathcal{E}(\rmI_\tau)$}\label{sec:mean_centering}

To justify our choice of regularizing $\mathcal{C}\bigl(\mathcal{E}(\rmI_\tau)\bigr)$ to follow the coefficient dynamics, we train the model with and without the mean-centering function $\mathcal{C}$.
Without $\mathcal{C}$, the learned representations consistently show larger latent norms in Fig.~\ref{fig:mean-centered}.
Since blurred image should always have smaller norms to satisfy Eq.~(\ref{eq:whippo_update2}), it results in mapping clear images to the latents with relatively larger norm magnitudes, inflating the value scales of the latent distribution.
Apparently, such an inflation pushes the model away from a generation-friendly latent distribution; Tab.~\ref{tab:mean-centered} shows that it results in a sharp decline in generation performance, as hinted from its dramatically high KL loss.

\subsection{Which channels correspond to low- and high-frequency components?}\label{appendix:channel_assignment}
In Sec.~\ref{sec:latent_structures} and Fig.~\ref{fig:channel_unveiling}, we reveal latent channels in a low-to-high frequency order in order to validate what information each channel encodes.
In this section, we describe how this ordering is determined.

Note that our regularization enforces the latent channels to mimic the dynamics of basis coefficients.
Since each basis coefficient captures a different frequency component and there is a bijective mapping between a latent channel and a basis, we can sort latent channels by the frequency order.
For example, let's say we designate a set of 2D Fourier basis functions: $\{\phi_n\}_{n=1}^{16} = \{\phi_{w,h}\}_{(w,h) \in \{0,1,2,3\} \times \{0,1,2,3\}}$.
Remark that each basis function is defined as follows:
\begin{equation}
    \phi_{w,h}(x,y) = \exp\Bigl(2\pi i \bigl(\frac{wx}{W} + \frac{hy}{H} \bigr)\Bigr).
\end{equation}
Here, a basis with smaller $w$ and $h$ corresponds to the basis that captures low frequency components.
Using this, we can group basis functions based on the frequency component they capture (low-to-high order)\footnote{
Technically, a magnitude of frequency $\phi_{w,h}$ captures is proportional to $w^2 + h^2$, but we group them by $w+h$ for simplicity.
}:
\begin{gather}\label{eq:basis_sort}
    \begin{split}
        \{\phi_{0,0}\}, \{\phi_{1,0}, \phi_{0,1}\}& , \{\phi_{2,0}, \phi_{1,1}, \phi_{0,2}\}, \\
        \{\phi_{3,0}, \phi_{2,1}, \phi_{1,2}, \phi_{0,3}\},
        \{\phi_{3,1}, &\phi_{2,2}, \phi_{1,3}\},
        \{\phi_{3,2}, \phi_{2,3}\},
        \{\phi_{3,3}\}.
    \end{split}
\end{gather}
Since we map $n$-th channel to resemble the $n$-th basis $\phi_n$, by the index mapping function $\iota$ (Eq.~(\ref{eq:index_mapping})), we can map each channel component to an element in Eq.~(\ref{eq:basis_sort}):
\begin{gather}\label{eq:channel_sort}
    \begin{split}
        \{\phi_{1}\}, \{\phi_{2}, \phi_{5}\}& , \{\phi_{3}, \phi_{6}, \phi_{9}\}, \\
        \{\phi_{4}, \phi_{7}, \phi_{10}, \phi_{13}\},
        \{\phi_{8}, &\phi_{11}, \phi_{14}\},
        \{\phi_{12}, \phi_{15}\},
        \{\phi_{16}\}.
    \end{split}
\end{gather}
Hence, we complete sorting channels in low-to-high frequency order.
We provide more intuitive illustrations to understand this channel-to-coefficient mapping in Fig.~\ref{fig:index_mapping}.
The grouping used for experiments in Sec.~\ref{sec:latent_structures} merges the first two groups and the last two groups to construct five groups: 
\begin{gather}
    \begin{split}
        g_1 = \{\phi_{1}, \phi_{2}, \phi_{5}\} ,& \quad  g_2 = \{\phi_{3}, \phi_{6}, \phi_{9}\}, \\
        g_3 = \{\phi_{4}, \phi_{7}, \phi_{10}, \phi_{13}\}, \quad
        g_4 = &\{\phi_{8}, \phi_{11}, \phi_{14}\}, \quad
        g_5 = \{\phi_{12}, \phi_{15}, \phi_{16}\}.
    \end{split}
\end{gather}

\input{figures/image_blurring_sequence}

\input{figures/index_mapping}

\subsection{Ablation studies}\label{appendix:ablation_studies}
We provide ablation studies across the type of state transition matrix $\mathbf{A}$ (Table~\ref{tab:A_ablation}), the regularization strength $\alpha$ (Table~\ref{tab:loss_ablation}), the maximum blur of an image $\tau_\mathrm{max}$ (Table~\ref{table:ablation_max_blur}), and the step size initialization value $\Delta$ (Table~\ref{table:ablation_step_size}).
Every ablation is conducted on the Flux tokenizer, using LightningDiT-B/2 as a generator.
Among all basis functions we implemented, using Fourier and Legendre bases leads to the best performance.
We stick to the Fourier $\mathbf{A}$ despite its worse generation performance compared to the Legendre $\mathbf{A}$, as we consider the Fourier's case provides better trade-off between reconstruction and generation, indicated by the lower rFID value.
Interestingly, random initialization of $\mathbf{A}$ leads to unstable convergence in learning, meaning that finding a properly structured matrix is important to implement SSM formulation to the regularization framework.
% Yet, all alternatives surpass the baseline, indicating the effectiveness of the proposed regularization.
We also ablated the regularization strength $\alpha$.
Note that we use $\alpha=0.25$ for all the experiments from the main manuscript (Tables~\ref{table:imagenet},~\ref{tab:mean-centered}).
As the regularization strength $\alpha$ increases, we observe both decrease in reconstruction and generation capability.
We hypothesize that increasing $\alpha$ over $0.25$ overly deteriorates the tokenizer's reconstruction ability and leads to the tokenizer being insufficient to properly decode the latent back to the pixel space.
For the maximum blur $\tau_\mathrm{max}$, the performance remains stable for $\tau_\mathrm{max} \in [8,10]$, while values outside of that range degrade generation.
Varying $\Delta$ over $[0.5, 2.0]$ changes performance only marginally, with a slight trade-off between reconstruction and generation.

\input{tables/ablations}

\subsection{Complexity analysis}\label{appendix:complexity_analysis}

We demonstrate the complexity of our regularizer in Table~\ref{table:complexity}.
Although our regularizer requires two feed-forward pass of the encoder, it barely affects the training efficiency of the tokenizer.

\input{tables/complexity_table}

\subsection{Application to different tasks}

We explore the broader applicability of structured state-space regularization, by broadening our experiments to following tasks:
\begin{itemize}
    \item \textbf{Text-conditioned generation} on CC12M~\citep{CC12M}, using the Flux tokenizer and a generator built with MMDiT~\citep{esser2024scaling} blocks.
    \item \textbf{Image denoising} on real-world denoising dataset SIDD~\citep{abdelhamed2018high}, based on the image restoration model Restormer~\citep{zamir2022restormer}.
\end{itemize}
For text-conditioned generation, we use Flux tokenizers fine-tuned on ImageNet with and without our regularizer, and train the corresponding models for 200K iterations, following the training recipe of MiniT2I~\citep{minit2i2026}.
For image denoising, we regularize the output of the refinement stage, the last Transformer block group before the final output convolution, with $\alpha=0.1$.

We report the results in Tables~\ref{table:cc12m}, ~\ref{table:sidd} and Figures~\ref{fig:cc12m},~\ref{fig:sidd}: applying the regularizer consistently improves performance on both tasks, highlighting the usefulness of frequency awareness in different computer vision tasks.

\input{tables/different_tasks}
\input{figures/cc12m_sidd}

\subsection{Additional qualitative examples}\label{appendix:additional_qualitative_comparison}

We provide additional qualitative examples of the channel revealing experiment (Fig.~\ref{fig:supp_channel_unveiling}) and the generation experiment from the Cosmos tokenizer (Fig.~\ref{fig:cosmos_qual1}).
We provide more examples from the Flux tokenizer in Fig.~\ref{fig:flux_qual1} as well.

\subsection{Societal impacts}~\label{appendix:societal_impacts}
Our work contributes to improving image tokenizers for generative modeling, which may positively impact applications requiring efficient and high-quality visual generation, including content creation, simulation, robotics, and scientific visualization. By improving latent-space organization without substantially increasing computational overhead, the proposed method may also help reduce training and inference costs for large-scale generative systems. However, as with other generative modeling technologies, improved image generation capabilities may also increase the risk of misuse, including the creation of deceptive or synthetic media.

\subsection{Limitations}\label{appendix:limitations}
One limitation of our work is that the proposed regularization primarily targets latent-space organization for generative modeling, and therefore may not consistently improve pixel-level reconstruction fidelity compared to objectives explicitly designed for reconstruction preservation. In addition, while our method demonstrates effectiveness across multiple tokenizer architectures, we currently evaluate it mainly on image generation settings, leaving its impact on broader downstream tasks such as representation learning, video generation, or multimodal modeling as future work.

\input{figures/supp_channel_reveal}

\clearpage

\input{figures/cosmos_generation}

\input{figures/flux_generation}

%% file: sections/2_related_work.tex
\subsection{Related work}
\label{sec:related_work}

\subsubsection{Tokenizers for diffusion models}
Modern generative models primarily employ the diffusion framework~\cite{ho2020denoising}, which models the distribution transformation from Gaussian noise to sample distribution~\cite{song2020score}.
A common practice in diffusion models is to employ tokenizers~\cite{rombach2022high} since performing the denoising process in the high-dimensional pixel space is prohibitively costly.
Specifically, a tokenizer is trained to map input data into a latent space and reconstruct them back, offering a compact low-dimensional representation of each sample that is utilized as input for diffusion models instead of raw visual data.
Examples of such a tokenizer include
VQVAE~\cite{vqgan}, SDVAE~\cite{rombach2022high}, Flux~\cite{Flux}, Cosmos~\cite{CosmosTokenizer}, and CogVideo~\cite{hong2022cogvideo}.

\subsubsection{Generation-friendliness of tokenizers}

Recent studies further explore desirable properties of autoencoders to enhance the diffusion process, such as enforcing rotation or scale equivariance between the pixel and latent spaces~\cite{EQVAE, skorokhodov2025improving}, constraining the latent space with fewer Gaussian mixture modes~\cite{chen2025masked}, or incorporating features from vision foundation models~\cite{vavae,repae,RAE}.
Our work also introduces a novel form of equivariance between the pixel and latent spaces that is effective for diffusion modeling, aligning Gaussian blurring in the pixel space with the corresponding operation in the 2D orthogonal basis coefficient space.

\subsubsection{State space models (SSMs)}

\noindent \textbf{1D SSMs}
SSMs have steadily evolved into a framework for sequence modeling, based on hidden state dynamics defined by a transition matrix. 
One of the pioneering studies on SSMs is HiPPO~\cite{gu2020hippo}, which constructs a compact representation of 1D sequences through projection onto orthogonal polynomials and derives the \textit{HiPPO matrix} $\mathbf{A}$, allowing the update of the compact representation that integrates new information in an online manner.
LSSL~\cite{gu2021combining} proposes a sequence-to-sequence layer based on the SSM formulation with the HiPPO matrix and establishes a grounding for its parameterization.
Subsequently, several approaches~\cite{gu2021efficiently, gupta2022diagonal, gu2022parameterization} have introduced ways to effectively parameterize the state transition matrix $\mathbf{A}$, while other stream of work has proposed different implementations of the SSM framework~\cite{smithsimplified, hasaniliquid, mehta2022long}.
The most well-known work is Mamba~\cite{gu2024mamba}, which introduces input-dependent SSM parameters with a selective scan mechanism.
It has given rise to numerous variants that apply SSM framework to various modalities~\cite{Vmamba, Vim, hu2024zigma, PointMamba, li2025videomamba}.

\noindent \textbf{Multi-dimensional SSMs}
Although SSMs are originally designed for 1D inputs, there have been several attempts to extend SSMs to multi-dimensional data.
While numerous studies have introduced sophisticated scanning algorithms that flatten the multi-dimensional input into 1D sequence to enable the application of SSMs~\cite{li2024mamba, zhang20252dmamba, Vim}, S4ND~\cite{nguyen2022s4nd} and 2D-SSM~\cite{baron20242}
take more theoretical approaches to derive the state-transition equation that works on $N$-dimensional inputs.
Specifically, they both designate a starting point (\eg, a corner of an image) and let the input tokens closer from the starting point are first integrated to the hidden state, building a multi-dimensional sequential process constructing a multi-dimensional input.
Our method also introduces a new SSM formulation that works on 2D inputs, but provides a different view on the input transformation: rather than building a sequence within a single image, the initial hidden state encodes the original input image and evolves to represent progressively blurred versions. 
We then regularize the hidden state updates based on this blurring transformation, enforcing them to follow the dynamics of basis functions—thereby endowing the hidden states with compressive properties.

%% file: figures/A_matrices.tex
\begin{figure}[htb]
    \centering
    \begin{subfigure}{0.45\textwidth}
        \centering
        \includegraphics[width=\linewidth]{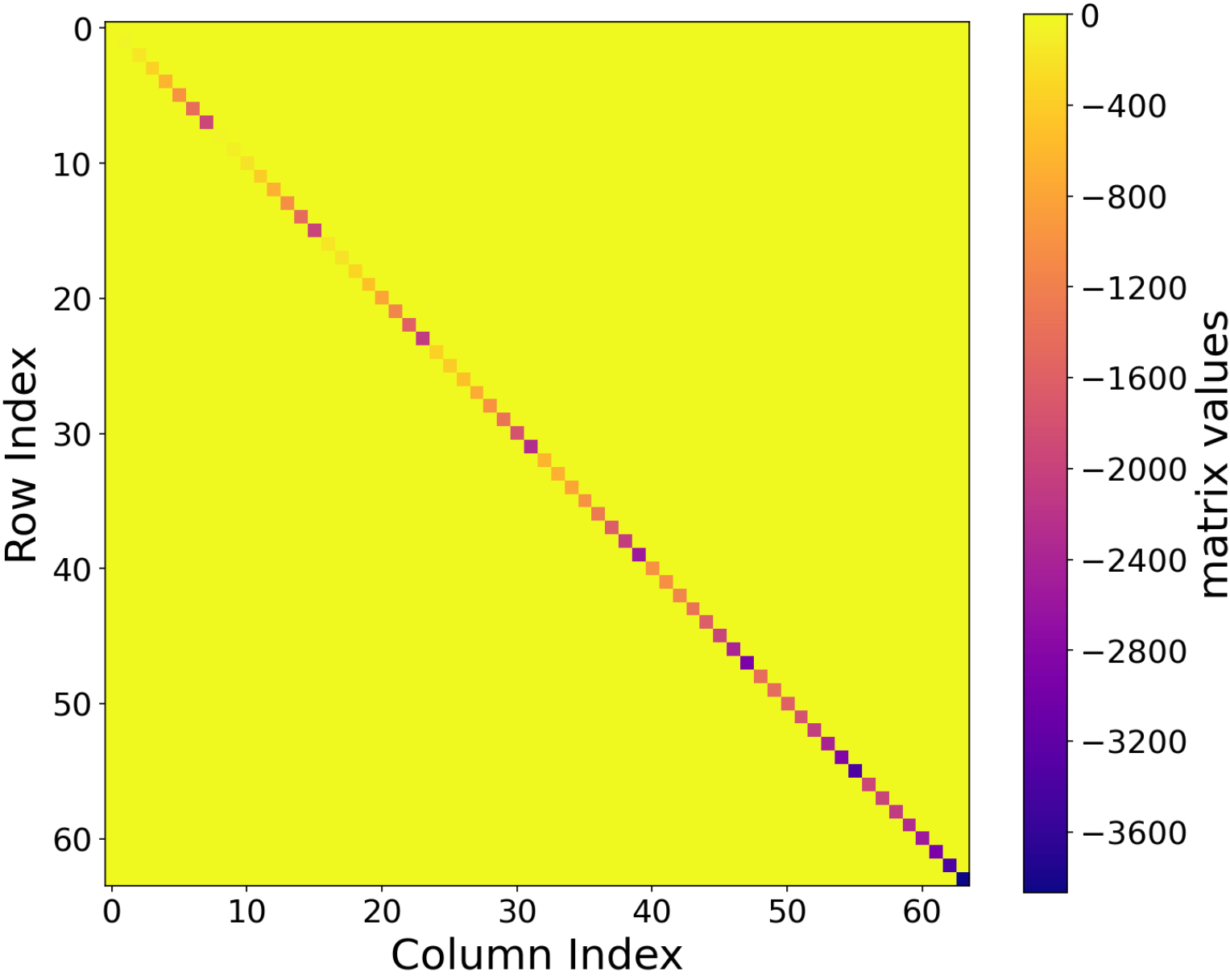}
        \caption{Visualization of Fourier $\mathbf{A}$.}
        \label{fig:img1}
    \end{subfigure}
    \hfill
    \begin{subfigure}{0.45\textwidth}
        \centering
        \includegraphics[width=\linewidth]{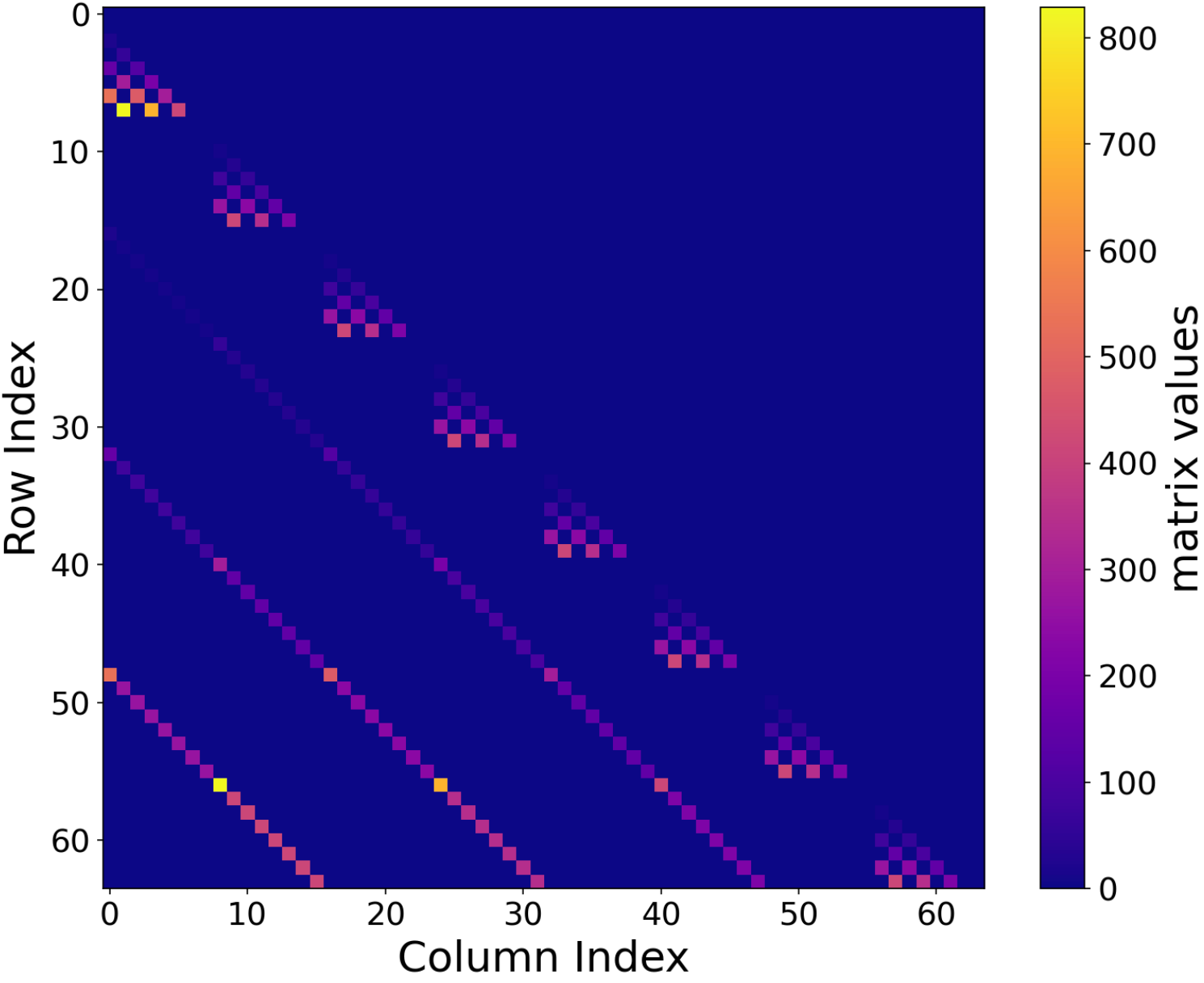}
        \caption{Visualization of Chebyshev $\mathbf{A}$.}
        \label{fig:img2}
    \end{subfigure}
    
    \vskip\baselineskip
    \begin{subfigure}{0.45\textwidth}
        \centering
        \includegraphics[width=\linewidth]{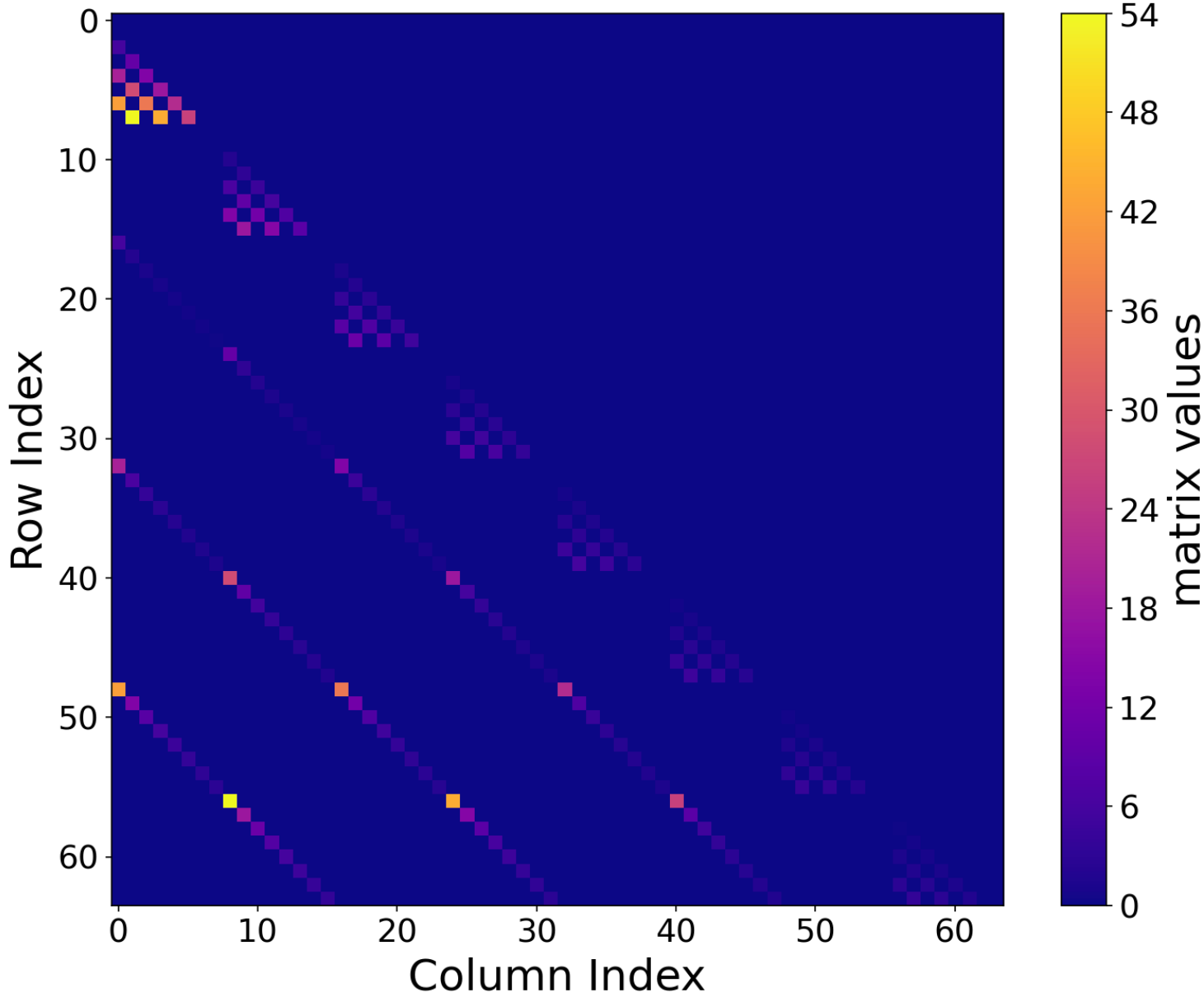}
        \caption{Visualization of Legendre $\mathbf{A}$.}
        \label{fig:img3}
        \vspace{6mm}
    \end{subfigure}
    \hfill
    \begin{subfigure}{0.45\textwidth}
        \centering
        \includegraphics[width=\linewidth]{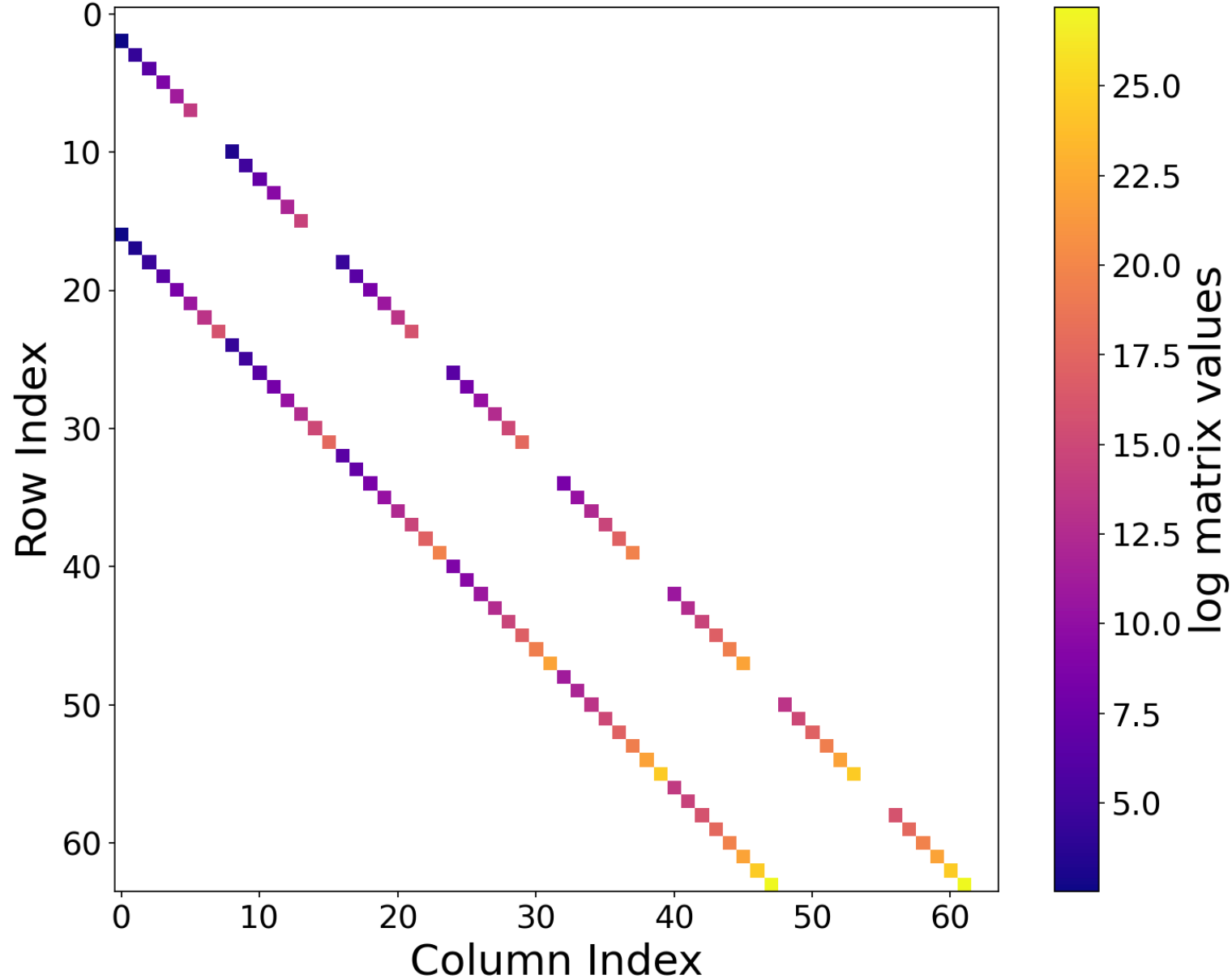}
        \caption{Visualization of Hermite $\mathbf{A}$. 
        Due to the exponential scale of matrix values, log values are displayed. Zero values marked white.}
        \label{fig:img4}
    \end{subfigure}
    
    \caption{\textbf{Visualization of derived} $\rmA$ \textbf{matrices.} 
    We set $H=W=8$. Every matrix we derived is very sparse, which enables efficient matrix-vector multiplication. To maintain training stability, we normalize the matrix to have maximum absolute value of 1.
    }
    \label{fig:A_matrices}
\end{figure}

%% file: tables/experiment_details1.tex
\begin{table*}[ht]
    \centering
    \vspace{3mm}
    \caption{
        Implementation details of the reconstruction experiments from Sec.~\ref{sec:experiments}.
    }\label{tab:experiment_details1}
    \vspace{-2mm}
    \scalebox{0.80}{
    \begin{tabular}{ccccccccccc}
    \toprule
     \multirow{2}{*}{Tokenizer} & {Num} & {batch} & {KL} & Warmup init. & {Base} & {Min} &  $\mathbf{A}, \Delta$ & Gradient\\
         & params & size & weight & learning rate & learning rate & learning rate & learning rate & clipping norm \\
    \midrule
     {Flux} & 83.8M & 32 & $10^{-6}$ & $1.0\times10^{-7}$ & $1.0\times10^{-5}$ & $1.0\times10^{-6}$ & $1.0\times10^{-6}$ & 50.0  \\
       {Cosmos} & 81.5M & 32 & {-} & $1.0\times10^{-7}$ & $1.0\times10^{-4}$ & $1.0\times10^{-5}$ & $1.0\times10^{-5}$ & 50.0 \\
    \bottomrule
    \end{tabular}
    }
    \vspace{-2mm}
\end{table*}

%% file: tables/experiment_details2.tex
\begin{table*}[ht]
    \centering
    \vspace{3mm}
    \caption{
        Hyperparameters of the regularization framework
    }\label{tab:whippo_implementation_details2}
    \vspace{-2mm}
    \scalebox{0.85}{
    \begin{tabular}{ccccccccccc}
    \toprule
     Tokenizer & {$\tau_\textrm{max}$} &{$\Delta_\mathbf{I}$} & {$\Delta$ init.} & Max $\mathbf{A}$ init. & $d_{\mathbf{z}}(\cdot, \cdot)$ & $d_{\mathbf{I}}(\cdot, \cdot)$ & $\lambda_\mathbf{z}$ & $\lambda_{\rmI} $ & $\lambda_\textrm{MAE}$ & $\lambda_\textrm{LPIPS}$\\
    \midrule
     Flux & 8.0 & 4.0 & 0.1 & 16.0 & MSE & MAE \& LPIPS & 0.0 & 1.0 & 1.0 & 0.1 \\
     Cosmos & 8.0 & 4.0 & 0.1 & 1.0 & MSE & MAE \& LPIPS & 0.05 & 1.0 & 1.0 & 0.1 \\
    \bottomrule
    \end{tabular}
    }
\end{table*}

%% file: figures/mean_centered.tex
\begin{figure*}[t!]
    \begin{subfigure}[b]{0.64\textwidth}
        \includegraphics[width=0.99\linewidth]{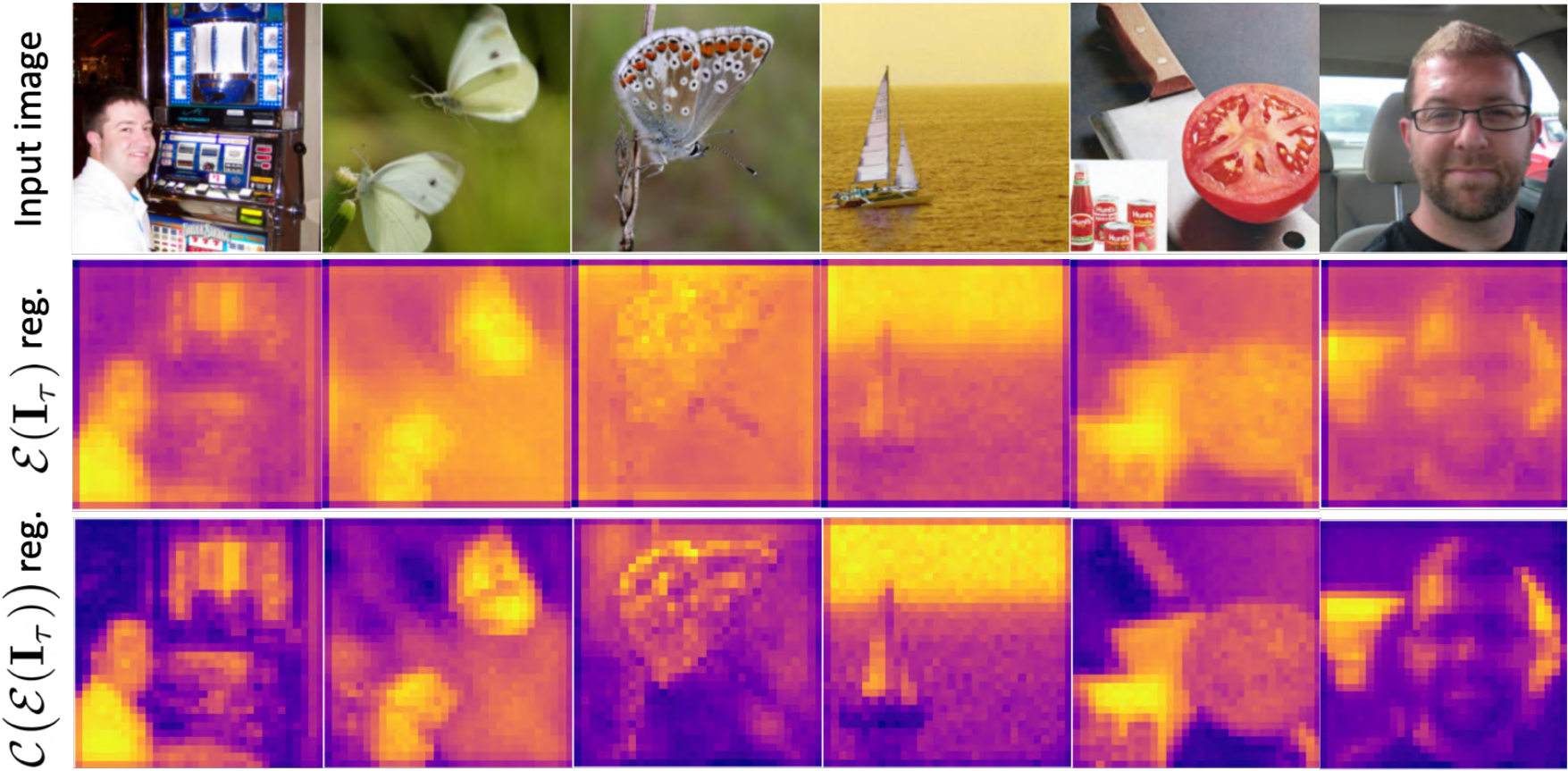}
        \captionsetup{format=hang, indention=0.0mm}
        \caption{
            Latent visualization, where channel dimension is reduced with $\ell_2$ norm.
            Brighter regions are higher values.
        }\label{fig:mean-centered}
    \end{subfigure}
    \hfill
    \begin{subfigure}[b]{0.34\textwidth}
        \centering
            \scalebox{0.85}{
            \begin{tabular}{lcc}
                \toprule
                  & $\mathcal{E}(\rmI_\tau)$ & $\mathcal{C}(\mathcal{E}(\rmI_\tau))$ \\
                \midrule
                PSNR $\uparrow$ & 31.53 & 31.54 \\
                LPIPS $\downarrow$ & 0.0373 & 0.0370 \\
                SSIM $\uparrow$   & 0.9083 & 0.9101  \\
                rFID $\downarrow$ & 0.849 &  0.901 \\
                KL $\downarrow$ &  10.45M & 203.2K \\
                gFID $\downarrow$   & 17.23 & 13.17  \\
                sFID $\downarrow$   & 10.02 & 6.71 \\
                IS $\uparrow$  & 88.14 &  105.99  \\
                precision $\uparrow$ & 0.74 & 0.71  \\
                recall $\uparrow$  & 0.31 & 0.48\\
                \bottomrule
            \end{tabular}
            }
        \captionsetup{format=hang, indention=0.0mm}
        \caption{
            Reconstruction and generation metric comparison
        }\label{tab:mean-centered}
    \end{subfigure}
    \caption{
        \textbf{Effect of spatial mean-centering function $\mathcal{C}.$}
        The one trained without $\mathcal{C}$ shows latents with higher latent norm, and exhibits tremendous degree of KL loss.
    }\label{fig:mean-centered-whole}
\end{figure*}

%% file: figures/image_blurring_sequence.tex
\begin{figure}[t!]
    \begin{center}
\includegraphics[width=1.0\linewidth]{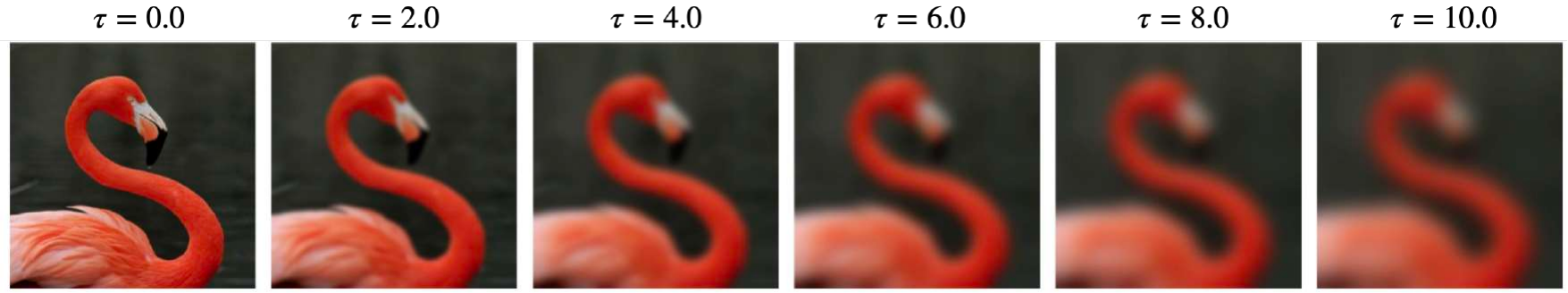}
    \end{center}

\caption{
Image displayed with respective $\tau$ values.
}
\label{fig:img_blurring_sequence}
\vspace{-2mm}
\end{figure}

%% file: figures/index_mapping.tex
\begin{figure}[t!]
    \begin{center}
\includegraphics[width=0.5\linewidth]{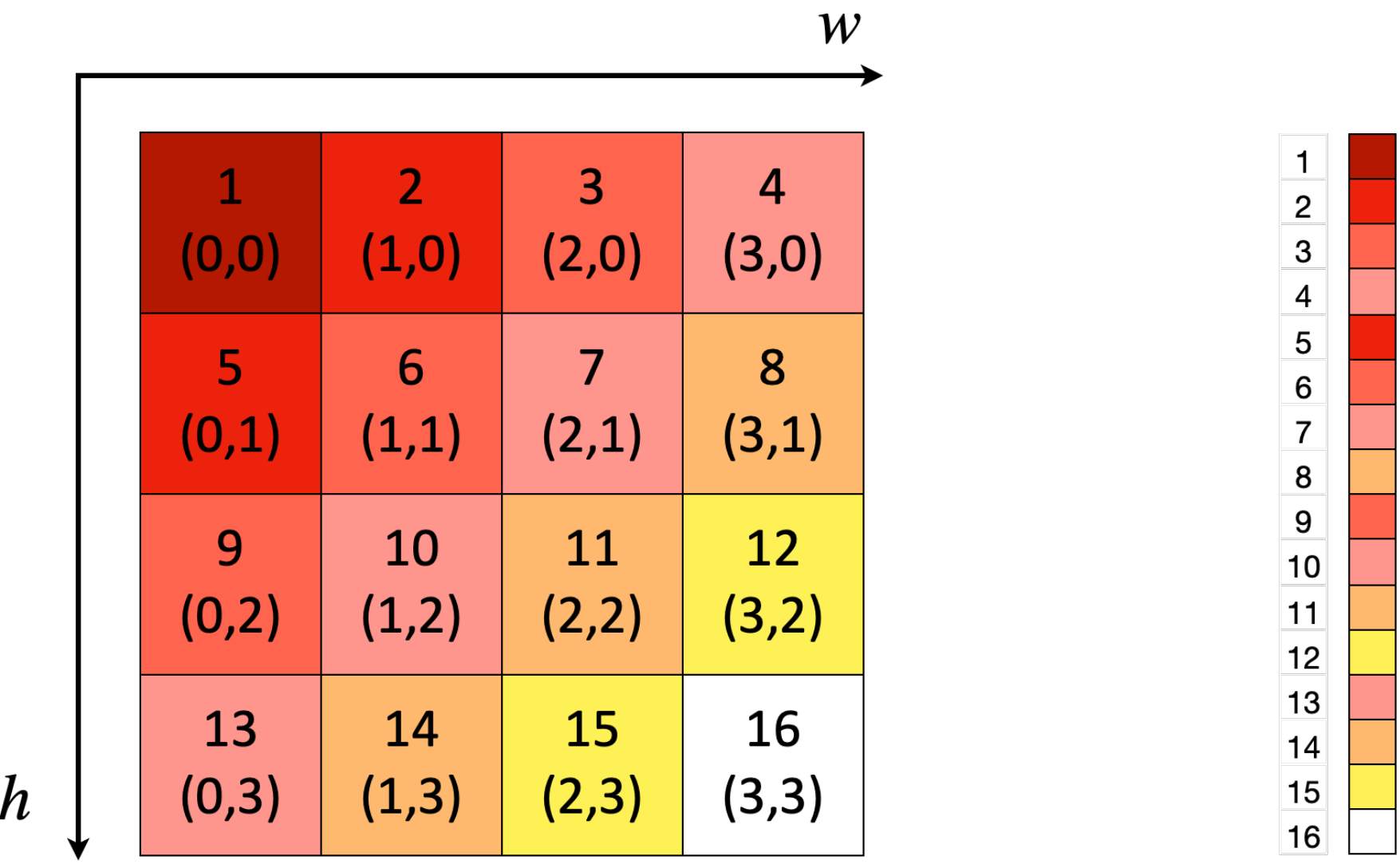}
    \end{center}
\caption{
(Left) Illustration of the channel-to-coefficient index mapping.
(Right) Channels illustrated with colors, where darker color indicates that a channel encodes lower frequency. 
}
\label{fig:index_mapping}
\end{figure}

%% file: tables/ablations.tex
\begin{table*}[t!]
    \caption{\textbf{Ablation studies.} We ablate across \textbf{(a)} the type of matrix $\mathbf{A}$, \textbf{(b)} the regularization strength $\alpha$, \textbf{(c)} the maximum blur $\tau_\mathrm{max}$, and \textbf{(d)} the step size $\Delta$.
    The rows containing the primarily chosen settings are highlighted.}
    \label{table:ablations}
    \hspace{-3mm}
    \begin{subtable}[t]{0.26\textwidth}
        \centering
        \caption{Type of $\mathbf{A}$}
        \setlength{\tabcolsep}{0.7pt}
        \input{tables/ablation1}
    \end{subtable}
    \hspace{-2mm}
    \begin{subtable}[t]{0.24\textwidth}
        \centering
        \caption{Regularization strength $\alpha$}
        \setlength{\tabcolsep}{0.7pt}
        \input{tables/ablation2}
    \end{subtable}
    \hspace{1mm}
    \begin{subtable}[t]{0.26\textwidth}
        \centering
        \caption{Maximum blur $\tau_{\mathrm{max}}$}
        \setlength{\tabcolsep}{0.7pt}
        \input{tables/ablation3}
    \end{subtable}
    \begin{subtable}[t]{0.19\textwidth}
        \centering
        \setlength{\tabcolsep}{0.7pt}
        \caption{Step size $\Delta$}
        \input{tables/ablation4}
    \end{subtable}
\vspace{-1mm}
\end{table*}

%% file: tables/ablation1.tex
\centering
\scalebox{0.57}{
\begin{tabular}{lcccc}
    \toprule
    $\mathbf{A}$ type & PSNR \(\uparrow\) & rFID \(\downarrow\) & gFID \(\downarrow\) & IS \(\uparrow\) \\
    \midrule
    \rowcolor{gray!20}
    Fourier   & {31.54} & {0.90} & 13.17 & 105.99 \\
    Legendre  & {31.57} & {0.97} & 12.61 & 108.84 \\
    Chebyshev & {31.56} & {0.90} & 13.21 & 105.18 \\
    Hermite   & {31.60} & {0.92} & 16.27 & 88.93 \\
    random    & \multicolumn{4}{c}{N/A (does not converge)} \\
    \bottomrule
\end{tabular}\label{tab:A_ablation}
}

%% file: tables/ablation2.tex
\centering
\scalebox{0.67}{
\begin{tabular}{lcccc}
    \toprule
    $\alpha$ & PSNR \(\uparrow\) & rFID \(\downarrow\) & gFID \(\downarrow\) & IS \(\uparrow\) \\
    \midrule
    0    & {31.70} & {0.87} & 16.08 & 89.32 \\
    \rowcolor{gray!20}
    0.25 & {31.54} & {0.90} & 13.17 & 105.99 \\
    0.5  & {30.95} & {1.01} & 13.73 & 111.65 \\
    0.75 & {29.83} & {1.28} & 15.48 & 112.02 \\
    \bottomrule
\end{tabular}\label{tab:loss_ablation}
}

%% file: tables/ablation3.tex
\centering
\scalebox{0.67}{
        \begin{tabular}{lcccc}
        \toprule
        $\tau_\mathrm{max}$ & PSNR $\uparrow$ & rFID $\downarrow$ & gFID $\downarrow$ & IS $\uparrow$ \\
        \midrule
         6 & 31.50 & 0.94 & 13.20 & 107.77 \\
         \rowcolor{gray!20}
         8 & 31.54 &0.90 & 13.17 & 105.99 \\
         10 & 31.54 & 0.93 & 13.02 & 106.88 \\
         12 & 31.53 & 0.92 & 13.96 & 101.48 \\
        \bottomrule
        \end{tabular}
        }\label{table:ablation_max_blur}

%% file: tables/ablation4.tex
\centering
\scalebox{0.67}{
    \begin{tabular}{lcccc}
    \toprule
    $\Delta$ & PSNR $\uparrow$ & rFID $\downarrow$ & gFID $\downarrow$ & IS $\uparrow$ \\
    \midrule
     0.1 & 31.60 & 0.91 & 14.06 & 100.54 \\
     0.5 & 31.50 & 0.93 & 13.04 & 101.34 \\
     \rowcolor{gray!20}
     1.0 & 31.54 & 0.90 & 13.17 & 105.99 \\
     2.0 & 31.50 & 0.96 & 13.00 & 106.86 \\
    \bottomrule
    \end{tabular}
    }    \label{table:ablation_step_size}

%% file: tables/complexity_table.tex
% \begin{table*}[t!]
% \caption{\textbf{Iteration speed and throughput of the proposed regularizer.} Measured with 4 RTX 6000 Ada GPUs, with regularization strength $\alpha=0.25$.}
% \centering
%     \scalebox{0.85}{
%     \begin{tabular}{lcccc}
%     \toprule
%     \multirow{2}{*}{Tokenizer} & speed & throughput & training time & memory \\
%      & (ms/it) & (img/s) & (h/epoch) & (GB) \\
%     \midrule
%      Flux & 338.7 & 94.3 & 3.77 & 13.09 \\
%      \rowcolor{gray!20}
%      \quad + Ours & 353.5 & 90.3 & 3.94 & 13.41 \\
%      Cosmos & 86.4 & 368.0 & 0.97 & 3.83 \\
%      \rowcolor{gray!20}
%      \quad + Ours & 91.9 & 345.5 & 1.03 & 4.12 \\
%     \bottomrule
%     \end{tabular}
%     }
%     \label{table:complexity}
% \end{table*}

% % preamble: \usepackage{multirow}, \usepackage[table]{xcolor}
% \newcommand{\rel}[1]{\textsubscript{\scriptsize\textcolor{gray}{#1\%}}}

\begin{table*}[t!]
\caption{\textbf{Training complexity of the proposed regularizer.} Measured with 4 RTX 6000 Ada GPUs, with regularization strength $\alpha=0.25$.
}
\centering
    \scalebox{0.85}{
    \begin{tabular}{lcccc}
    \toprule
    \multirow{2}{*}{Tokenizer} & speed & throughput & training time & memory \\
     & (ms/it) & (img/s) & (h/epoch) & (GB) \\
    \midrule
     Flux & 338.7\norel & 94.3\norel & 3.77\norel & 13.09\norel \\
     \rowcolor{gray!20}
     \quad + Ours & 353.5\rel{+4.4} & 90.3\rel{$-$4.2} & 3.94\rel{+4.4} & 13.41\rel{+2.4} \\
     Cosmos & 86.4\norel & 368.0\norel & 0.97\norel & 3.83\norel \\
     \rowcolor{gray!20}
     \quad + Ours & 91.9\rel{+6.4} & 345.5\rel{$-$6.1} & 1.03\rel{+6.4} & 4.12\rel{+7.6} \\
    \bottomrule
    \end{tabular}
    }
    \label{table:complexity}
    \vspace{-2mm}
\end{table*}

%% file: tables/different_tasks.tex
\begin{table}[t!]
    \begin{minipage}{0.53\textwidth}
    \caption{Text-conditioned generation results on CC12M~\citep{CC12M}, using CLIP~\citep{CLIP} as the text encoder}
    \vspace{2mm}
    \centering
    \setlength{\tabcolsep}{6.6pt}
    \scalebox{0.9}{
    \begin{tabular}{lcc}
    \toprule
    {Tokenizer} & COCO FID $\downarrow$ & CLIP score $\uparrow$ \\
    \midrule
     Flux & {28.57} & {0.290} \\
     \rowcolor{gray!20}
     Flux + Ours & \textbf{24.90} & \textbf{0.300} \\
    \bottomrule
    \end{tabular}
    }
    \label{table:cc12m}
    \end{minipage}
    \hfill
    \begin{minipage}{0.43\textwidth}
    \caption{Image denoising results on SIDD~\citep{abdelhamed2018high}, based on Restormer~\citep{zamir2022restormer}}
    \vspace{2mm}
    \centering
    \scalebox{0.9}{
    \begin{tabular}{lcc}
    \toprule
    Model & PSNR $\uparrow$ & SSIM $\uparrow$ \\
    \midrule
     Restormer & {39.5803} & {0.9111} \\
     \rowcolor{gray!20}
     Restormer + Ours & \textbf{39.8939} & \textbf{0.9128} \\
    \bottomrule
    \end{tabular}
    }
    \label{table:sidd}
    \end{minipage}
\end{table}

%% file: figures/cc12m_sidd.tex
\begin{figure}[t!]
    \begin{minipage}{0.35\textwidth}
    \centering
    \includegraphics[width=\linewidth]{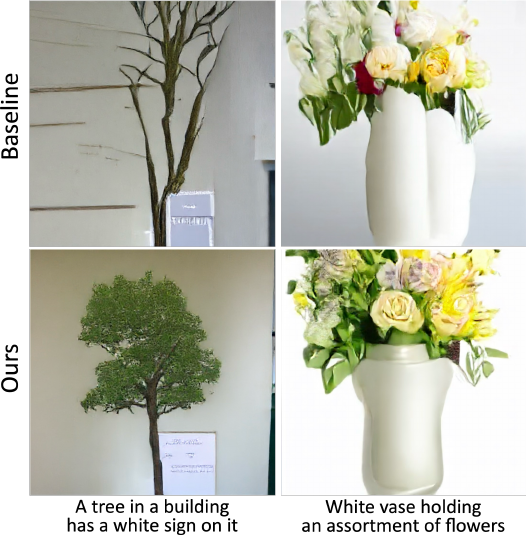}
    \caption{Qualitative examples from text-conditioned generation}\label{fig:cc12m}
    \end{minipage}
    \hfill
    \begin{minipage}{0.62\textwidth}
    \centering
    \includegraphics[width=\linewidth]{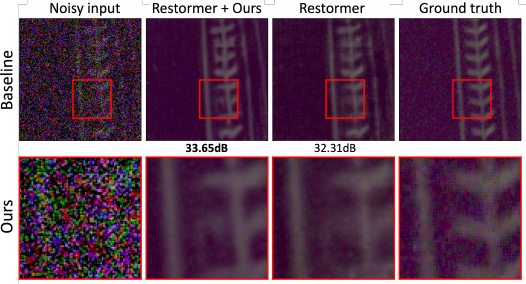}
    \caption{Qualitative examples from image denoising}\label{fig:sidd}
    \end{minipage}
    \vspace{-2mm}
\end{figure}

%% file: figures/supp_channel_reveal.tex
\begin{figure*}[t!]
    \centering
    \includegraphics[width=1.0\linewidth]{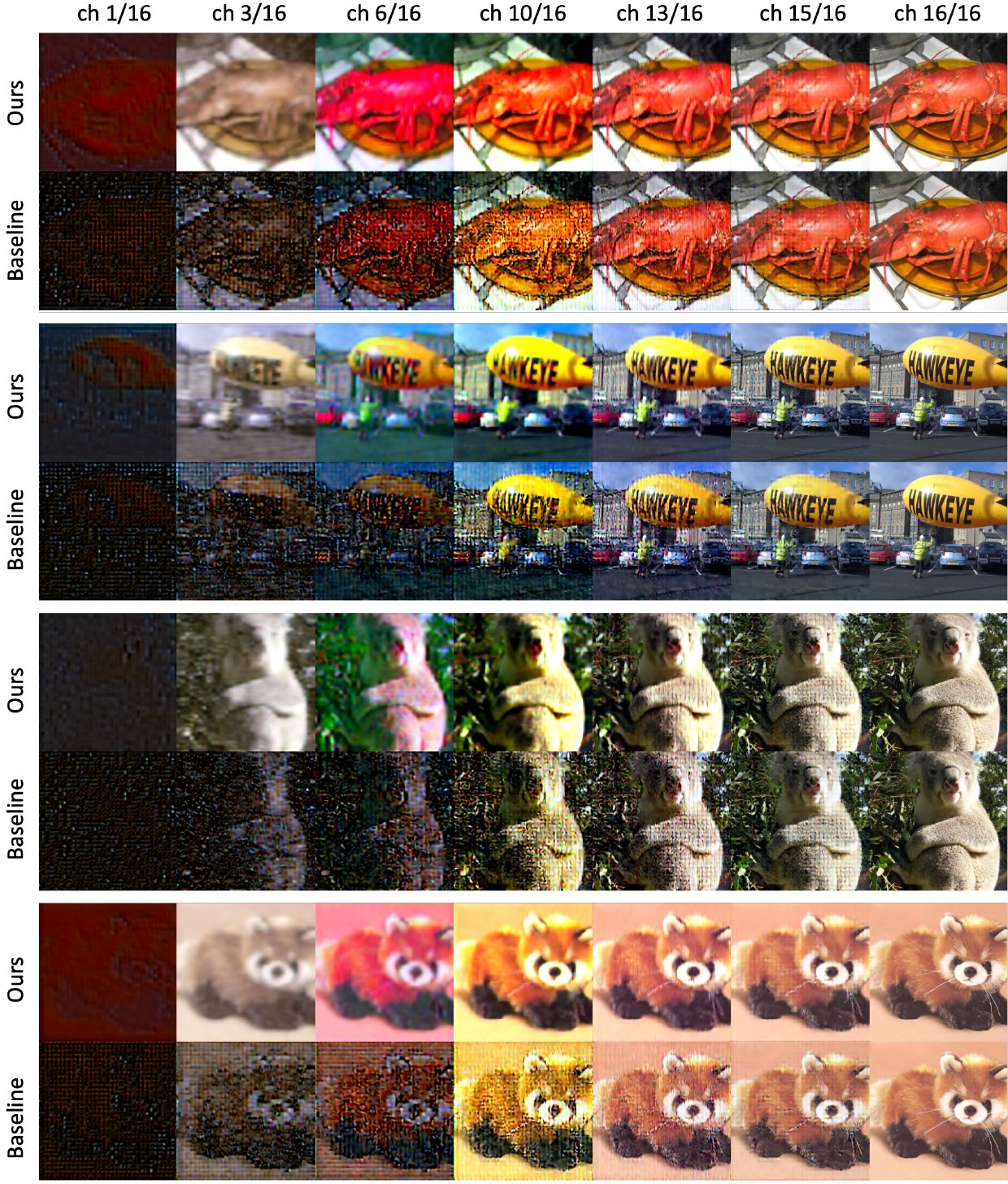}

    \vspace{2mm}
    
    \caption{Additional results from the channel revealing experiment}
    \label{fig:supp_channel_unveiling}
\end{figure*}

%% file: figures/cosmos_generation.tex
\begin{figure*}[t!]
    \centering
    \includegraphics[width=\linewidth]{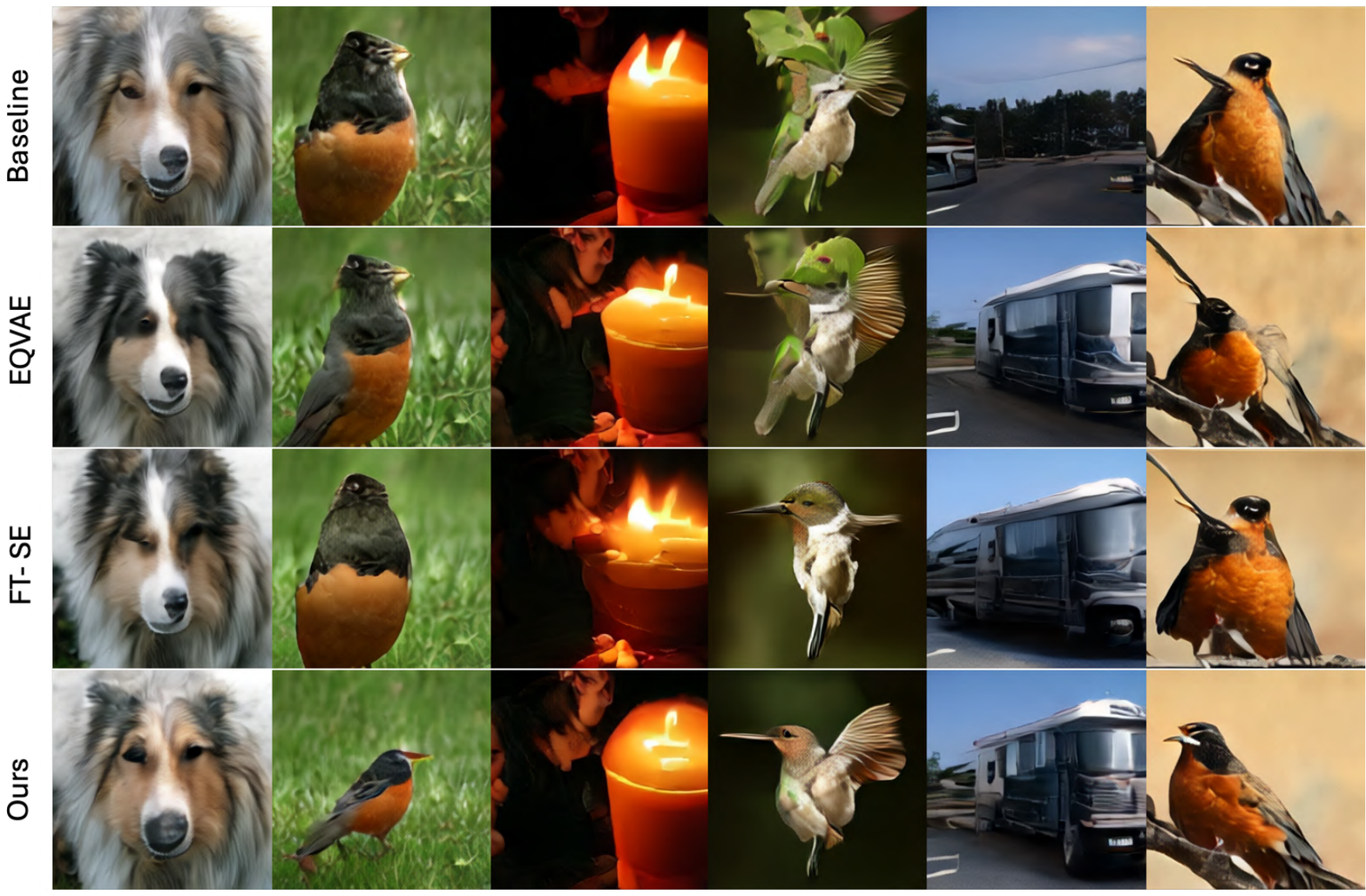}

    \vspace{7mm}

    \includegraphics[width=\linewidth]{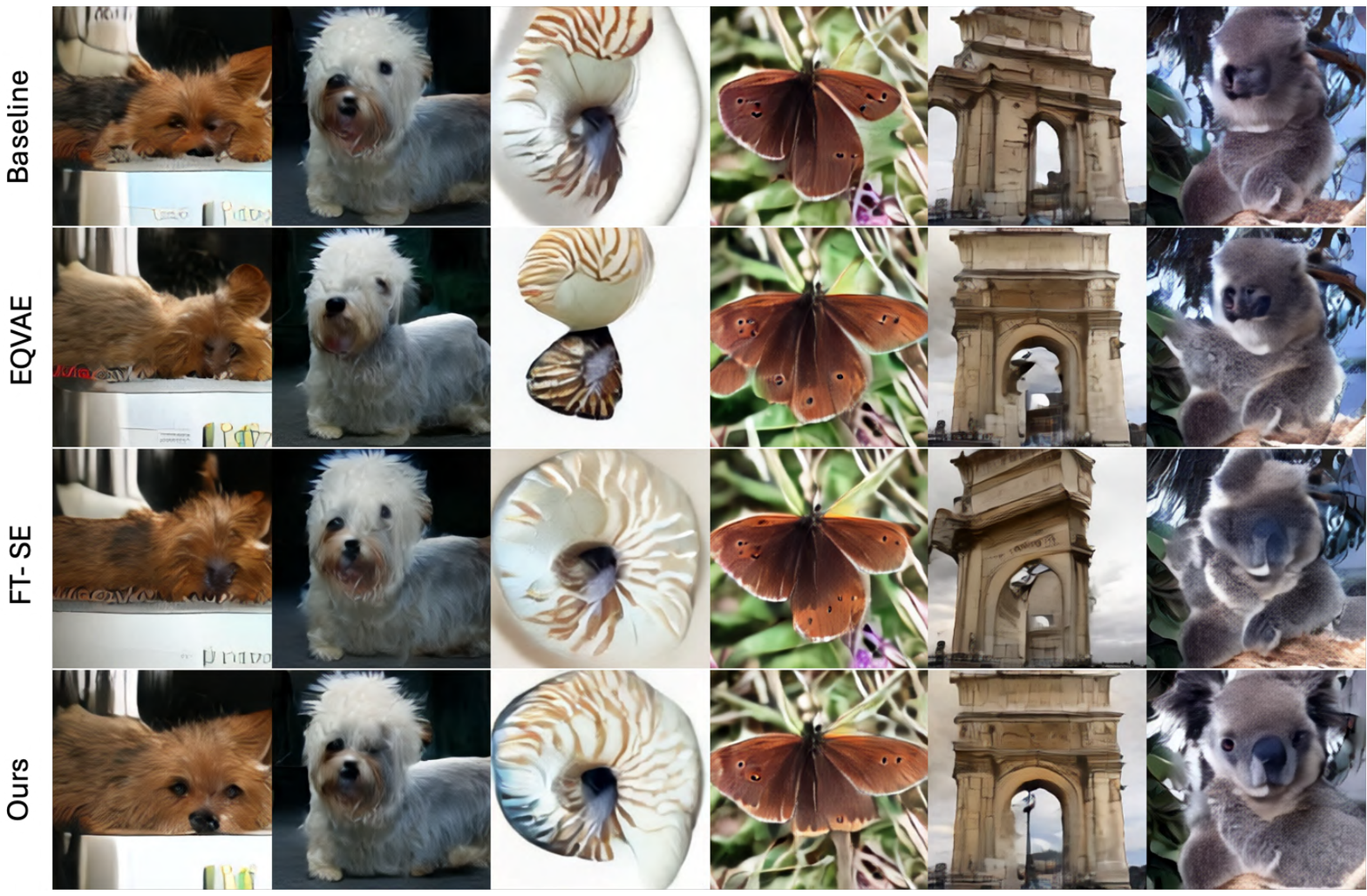}
    
    \vspace{2mm}
    
    \caption{Generation results from the Cosmos tokenizer}
    \label{fig:cosmos_qual1}
\end{figure*}

%% file: figures/flux_generation.tex
\begin{figure*}[t!]
    \centering
    \includegraphics[width=\linewidth]{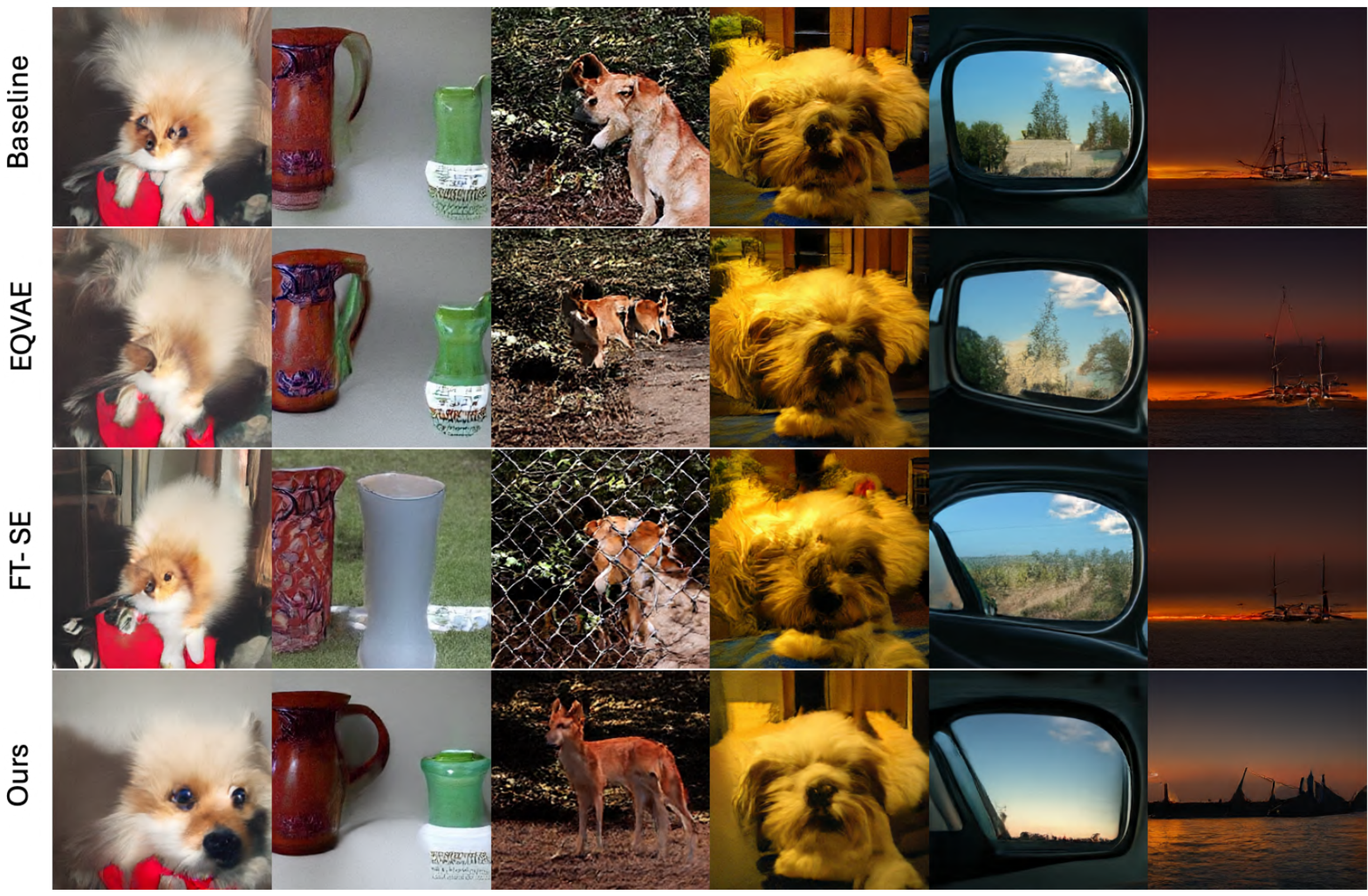}

    \vspace{4mm}
    \includegraphics[width=\linewidth]{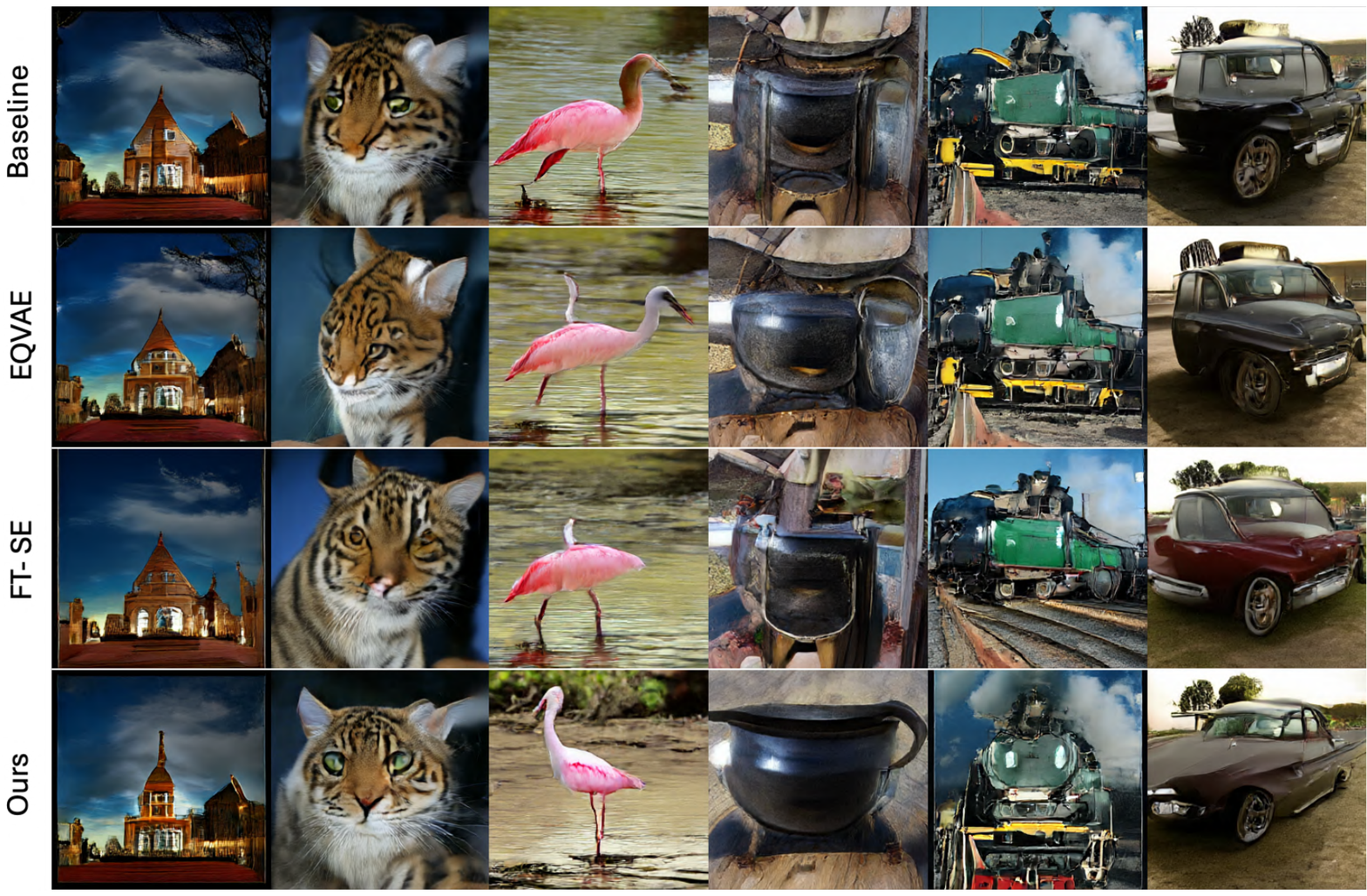}

    \vspace{2mm}
    
    \caption{Additional generation results from the Flux tokenizer}
    \label{fig:flux_qual1}
\end{figure*}